\documentclass[lettersize,journal]{IEEEtran}
\usepackage{amsmath,amsfonts}
\usepackage{algorithmic}
\usepackage{algorithm}
\usepackage{array}
\usepackage{tabularx} 
\usepackage{adjustbox}
\usepackage{url}              %
\usepackage{color}
\usepackage{subcaption}
\usepackage{stfloats}
\usepackage{url}
\usepackage{verbatim}
\usepackage{textcomp}
\usepackage{soul}
\usepackage{graphicx}
\usepackage{cite}
\usepackage{xcolor}
\usepackage{multirow}
\usepackage{pifont}
\usepackage{mathabx}
\usepackage{booktabs}%

\def\BibTeX{{\rm B\kern-.05em{\sc i\kern-.025em b}\kern-.08em
    T\kern-.1667em\lower.7ex\hbox{E}\kern-.125emX}}
\usepackage{balance}

\begin{document}

\title{Pedestrian Attribute Editing for \\Gait Recognition and Anonymization}

\author{Jingzhe Ma{$^\dag$}, Dingqiang Ye{$^\dag$}, Chao Fan{$^\dag$}, and Shiqi Yu
\thanks{Jingzhe Ma, Dingqiang Ye, Chao Fan, and Shiqi Yu are both with the Department of Computer Science and Engineering, Southern University of Science and Technology, Shenzhen 518055, China. E-mail: \{12031127, 12232422, 12131100\}@mail.sustech.edu.cn, and yusq@sustech.edu.cn}
\thanks{$\dag$ means that these authors contributed equally to this work. }
\thanks{Corresponding author: Shiqi Yu.}
}


\markboth{IEEE Transactions on Pattern Analysis and Machine Intelligence,~Vol.~XX, No.~YY, Month~2024}%
{Ma \MakeLowercase{\textit{et al.}}: Pedestrian Attribute Editing for Gait Recognition and Anonymization}

\maketitle

\begin{abstract}
As a kind of biometrics, the gait information of pedestrians has attracted widespread attention from both industry and academia since it can be acquired from long distances without the cooperation of targets.
In recent literature, this line of research has brought exciting chances along with alarming challenges: On the positive side, gait recognition used for security applications such as suspect retrieval and safety checks is becoming more and more promising. 
On the negative side, the misuse of gait information may lead to privacy concerns, as lawbreakers can track subjects of interest using gait characteristics even under face-masked and clothes-changed scenarios. 
To handle this double-edged sword, we propose a gait attribute editing framework termed \textbf{GaitEditor}. 
It can perform various degrees of attribute edits on real gait sequences while maintaining the visual authenticity, respectively used for gait data augmentation and de-identification, thereby adaptively enhancing or degrading gait recognition performance according to users' intentions.
Experimentally, we conduct a comprehensive evaluation under both gait recognition and anonymization protocols on three widely used gait benchmarks.
Numerous results illustrate that the adaptable utilization of GaitEditor efficiently improves gait recognition performance and generates vivid visualizations with de-identification to protect human privacy.
To the best of our knowledge, GaitEditor is the first framework capable of editing multiple gait attributes while simultaneously benefiting gait recognition and gait anonymization.
The source code of GaitEditor will be available at https://github.com/ShiqiYu/OpenGait.
\end{abstract}


\begin{figure*}[t]
    \centering
    \includegraphics[width=0.95\linewidth]{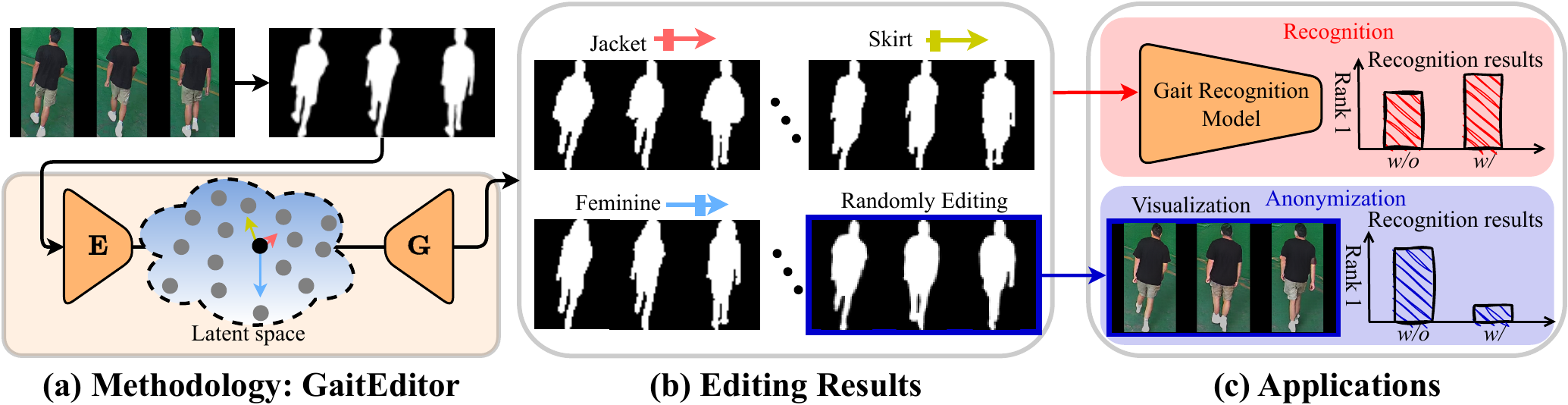}
    \caption{We present \textbf{GaitEditor} (sub-figure (a)) to enable attribute-conditioned semantic edits on real gait sequences (sub-figure (b)), thereby boosting the performance of gait recognition, all while mitigating privacy concerns (sub-figure (c)). For \textbf{GaitEditor}, we project the real gait data into the built latent space by well-designed encoder $\mathbf{E}$, manipulate latent codes in various semantic directions and project the altered ones to the image domain via the generator $\mathbf{G}$.} 
    \label{fig:fig1_overview}
\end{figure*}

\section{Introduction}\label{sec1}

\IEEEPARstart{N}{\textit{othing}} \textit{is perfect}. 
Pedestrian gait, characterized by the dynamic displacement of the body's center of gravity, encompasses the synchronized movement of the lower limbs and trunk to facilitate locomotion~\cite{alharthi2019deep}.
Notably, the gait information requires no target coordination and can be easily acquired from a long distance, rendering its applications double-edged.
On the positive side, gait serves as a biometric modality for individual identification, specifically in the field of gait recognition~\cite{shen2022comprehensive}.
This capability plays a critical role in crime investigation, surveillance systems, and social security~\cite{bouchrika2011using, iwama2013gait, lynnerup2014gait, wu2016comprehensive}.
On the downside, the utilization of gait information can potentially expose pedestrians' personal details, thereby raising privacy concerns. 
Even if the face is masked/swapped, and clothing is changed during walking, the unique characteristics inherent in a pedestrian's gait can still be identified and tracked.

Recently, with the rise of deep generative methods~\cite{kingma2013auto, goodfellow2014generative, karras2019style, ho2020denoising}, particularly Style-based Generative Adversarial Networks (GANs)~\cite{karras2019style, karras2020analyzing, karras2020training, karras2021alias}, object attribute editing has continuously attracted increasing interests.
Within this broader context, pedestrian gait attribute editing, referred to as gait attribute editing for convenience, emerges as a unique subfield, focusing on the manipulation of the bodily appearance (spatial), motion (temporal) or both aspects of pedestrians in walking. 
Several works have demonstrated that the gait attribute editing methodology can benefit recognition~\cite{yu2017gaitgan,li2020gait,chen2021multi, zhang2023large} or de-identification~\cite{ma2022identity, hirose2022anonymization}, \textit{i.e.}, \textbf{merely attending either the positive or negative side}.
In the paper, we aim to design a unified method for gait attribute editing that holds the promise of deftly navigating this double-edged sword.


In the field of gait recognition, a major challenge lies in addressing various factors that may drastically affect gait appearances~\cite{wu2016comprehensive}, such as cross-clothing, cross-viewpoints, and cross-carrying challenges.
Thus, many works~\cite{yu2017gaitgan,li2020gait,chen2021multi, zhang2023large} have attempted to tackle these challenges by editing or generating gait attributes using generative methods to improve the performance of gait recognition.
Although these methods have made these factors diverse and generated vivid silhouettes, they either rely heavily on paired data samples during training~\cite{yu2017gaitgan,li2020gait,chen2021multi}, or they create virtual gait datasets with fixed scales~\cite{zhang2023large}. In the field of gait anonymization, several studies have been proposed with a primary objective: to prevent gait recognition models from identifying individuals while minimizing modification to their overall texture appearance~\cite{hirose2022anonymization}.
Typically, these developments involve manipulating individual-specific factors, such as the static body shape features~\cite{agrawal2011person, mitsugami2010privacy, tieu2017approach}, dynamic motion features, or both~\cite{ma2022identity, hirose2022anonymization}.
However, these methods tend to edit the body shape and walking phase with homogeneous appearances for different people, which may lead to anonymized gait sequences being potentially reversible by deep learning models.
For reliable gait identity protection, it is also necessary to consider the diversity of variations of gait attributes.


The above discussion about gait recognition and anonymization reflects the main problem of these methods, that is, they could only edit restricted attributes. 
In this paper, by leveraging a latent space of style-based GANs and controlling the degree of manipulation in various semantic directions, we find that the diversity of gait attributes can be improved. 
Such advancement is vital for both augmenting gait datasets and achieving pedestrian anonymity, thereby ultimately achieving two purposes: improving recognition accuracy and preserving individual privacy. 
Notably, our focus is primarily on editing the body appearance of pedestrians.

Different from previous using style-based GANs for face editing tasks~\cite{abdal2019image2stylegan,shen2020interpreting, abdal2020image2stylegan++, wu2021stylespace, moon2022interestyle}, \textbf{four key challenges} inherent to gait need to be considered:
a) Binary gait silhouettes contain only sparse structural information, such as edges and shapes, which may make it hard to disentangle semantically controllable gait attributes. 
b) The gait data is sequential and not just an image, so editing gait sequences necessitates the maintenance of temporal continuity and identity consistency.
c) Posture changes usually damage the fidelity of face and object when conducting face editing~\cite{fruhstuck2023vive3d} and object editing~\cite{liu2023delving}, respectively.
Therefore, editing gait when crossing a viewpoint may be quite a challenging task.
d) The different objectives of gait recognition and anonymization raise an additional challenge: how to effectively apply a designed unified framework to these two distinct tasks.

To overcome these challenges, we propose a pioneering gait attribute editing methodology, named \textbf{GaitEditor}, illustrated in Fig.~\ref{fig:fig1_overview} (a). 
Firstly, a style-based GAN network is trained to transform random noise codes into photorealistic silhouettes, thereby establishing a semantically rich latent space. 
Within this space, various semantically controllable directions can be discovered.
We then apply GAN inversion~\cite{xia2022gan} to project real gait sequences into this latent space by an encoder, while maintaining temporal and identity integrity. 
Specifically, a spatial-temporal mix block and a carefully designed identity loss are integrated to safeguard these aspects. 
Manipulation of the latent codes along these directions enables the editing of corresponding attributes of the input gait sequences. 
Furthermore, we introduce a viewpoint translation branch in training to address cross-view editing challenges.


Experimentally, we conduct a user study to systematically analyze the authenticity of the edited attributes.
The results demonstrate that GaitEditor can edit various gait attributes, \textit{e.g.}, shirt, pants, viewpoint, body size, age-like, and gender-like, as shown in Fig.~\ref{fig:fig1_overview} (b).
Additionally, the straightforward design of GaitEditor allows it to function as a plug-and-play module, easily integrated into downstream tasks, as illustrated in Fig.\ref{fig:fig1_overview}(c).
Benefiting from this convenient integration, we observe that cleverly controlling the strength and frequency of edits, named the degree of edits, affects the consistency of identity before and after editing.
Given input gait sequences, minor and subtle edits bring minimal loss of identity consistency and are suitable for gait data augmentation.
Conversely, intense and substantial edits may weaken identity consistency, but they still yield vivid visualization in both silhouette and RGB format.
These insights significantly facilitate the application of GaitEditor to gait recognition and anonymization.
Overall, this paper makes the following contributions.


\begin{itemize}
    \item To our knowledge, GaitEditor represents the first attempt that edits various gait attributes in an unsupervised manner using a single straightforward generator, significantly facilitating the practicability of gait editing research. 
    \item We successfully apply GaitEditor as an online data augmentation module to enhance gait recognition algorithms. On the challenging cross-clothing dataset CCPG, notable improvements are observed across several popular gait models. Specifically, GaitSet, GaitPart, GaitBase, and DeepGaitV2 respectively exhibited improvements in average rank-1 accuracy by +1.00\%, +2.12\%, +1.60\%, and +1.29\% under various walking conditions.
    \item For Gait anonymization, GaitEditor can anonymize the identity within a gait sequence while maintaining the natural texture appearance information. Under the GaitBase framework, GaitEditor significantly disrupted rank-1 accuracy by -89.24\%, -57.34\%, and -56.00\% on three challenging datasets: CCPG, OU-MVLP, and Gait3D, illustrating its potent capability to safeguard personal identity in gait data.
\end{itemize}

\section{Related Works}\label{sec2}

\subsection{Gait Recognition}

Existing gait recognition methods can be generally grouped into two distinct categories: model-based and appearance-based.
Among them, the former utilizes the 2D/3D underlying structure of the human body, such as the skeleton and SMPL model, as input to infer the identity information~\cite{ariyanto2011model, liao2020model, teepe2021gaitgraph, teepe2022towards, li2022multi, fu2023gpgait, li2021end}.
Mostly, these body models tend to explicitly exclude gait-unrelated visual factors like carrying and dressing items. 
During this process, one of the most discriminative gait characteristics, i.e., the shape of the human body, is usually damaged or even removed simultaneously. 
This is considered the main cause of unsatisfactory recognition performance achieved by model-based methods compared with the appearance-based ones which focus more on gait appearance~\cite{fan2023opengait, zheng2022gait}.

Appearance-based gait recognition methods~\cite{chao2021gaitset, fan2020gaitpart, lin2021gait, fan2023opengait, fan2023exploring, liang2022gaitedge} mostly extract gait features from silhouette or RGB images, thus shape features are easily captured.
These methods are commonly focused on spatial feature extraction and gait temporal modeling.
In particular, GaitSet~\cite{chao2021gaitset} is one of the most renowned methods because of its novel idea that regards gait sequence as an unordered set.
GaitPart~\cite{fan2020gaitpart} meticulously explores the local details of input silhouettes and models temporal dependencies using the Micro-motion Capture Module~\cite{fan2020gaitpart}.
GaitGL~\cite{lin2021gait} argues that spatially global gait representations often overlook important details, while local region-based descriptors fail to capture relationships among neighboring parts. Consequently, GaitGL introduces global and local convolution layers~\cite{lin2021gait}.
OpenGait~\cite{fan2023opengait} revisited several representative works from an experimental perspective and constructed a robust baseline model, GaitBase. 
More recently, DeepGaitV2~\cite{fan2023exploring} presents a unified perspective to explore how to construct deep models for state-of-the-art gait recognition, bringing a breakthrough improvement on several challenging benchmarks.

\subsection{Gait Anonymization}
Gait anonymization methods, categorized into visual abstraction and replacement~\cite{hirose2022anonymization}, achieve anonymization by either pixelizing or blurring relevant human regions~\cite{agrawal2011person}, or subtly deforming silhouettes to prevent successful identification by gait recognition models.
There is limited research on visual abstraction-based gait anonymization. 
Specifically, Agrawal \textit{et al.}~\cite{agrawal2011person} introduced a technique involving the application of blurring filters to the entire human area in a video for gait anonymization. 
Mitsugami \textit{et al.}~\cite{mitsugami2010privacy} recommended the use of rod-like symbols placed over individual human regions in surveillance videos as a measure to protect individuals' privacy.
However, visual abstraction-based methods often result in an unnatural appearance.
To address this issue, Tieu \textit{et al.} introduced replacement-based methods for gait anonymization in their a serial of works~\cite{tieu2017approach, tieu2019spatio, tieu2019rgb}.
They specifically utilize generative methods to synthesize a new silhouette, slightly different from the target silhouette, by combining the target silhouette and a noise silhouette.
Furthermore, Hirose \textit{et al.}~\cite{hirose2022anonymization} deformed each frame's silhouette in a gait sequence from both the body shape and the walking phase to prevent accurate recognition of the person in the input video.
However, these methods tend to edit the body shape and walking phase with homogeneous appearances for different people, which may lead to anonymized gait sequences being potentially reversible by deep learning models.
In contrast, GaitEditor can autonomously and repeatedly edit various attributes, thereby ensuring the diversity of gait attributes.

\subsection{Gait Data Synthesis}

There are many methods leveraging the popular generative models to produce vivid gait data. 
For example, Yu \textit{et al.} in~\cite{yu2017gaitgan} produced the view-invariant Gait Energy Images (GEIs)~\cite{han2005individual} via Generative Adversarial Networks (GANs).
They used an identity discriminator to ensure the identity consistency of generated GEIs. 
Li \textit{et al.}~\cite{li2020gait} disentangled specific covariate attributes in latent space by contrast learning for synthesizing new GEIs, thereby transferring the covariate feature from one subject to the target. 
Chen \textit{et al.}~\cite{chen2021multi} presented a Multi-view Gait Generative Adversarial Network (MvGGAN) to generate gait sequences with different viewpoints, handling the lack of viewpoints under existing gait datasets.
Xiang \textit{et al.}~\cite{xiang2022multi} introduced the prior knowledge of person re-identification and gait recognition into the training loss to preserve identity consistency when transferring a subject viewpoint.
Recently, Zhang \textit{et al.}~\cite{zhang2023large} leveraged the Unity3D toolkit to build a large-scale virtual gait dataset, thereby diversifying the identity-unrelated factors of gait datasets.
Although these methods obtain vivid gait data, they require labeled data pairs for training or aim to build a fixed-scale virtual gait dataset~\cite{zhang2023large} by operating a single, specific gait attribute, e.g., the camera viewpoints~\cite{chen2021multi, xiang2022multi} or carrying changes~\cite{li2020gait}. 
In contrast, GaitEditor can act as an online module to augment the input gait sequences with various gait attribute changes.

\subsection{Style-based GANs and GAN Inversion}
Style-based GANs, \textit{i.e.} StyleGAN~\cite{karras2019style} refer to generating images via implicitly learned hierarchical latent styles.
In particular, this model utilizes the inputs of style latent code (obtained by a mapping network) and stochastic variation (achieved by noise layers) for image synthesis. 
In order to improve the perceptual quality of generated images, the StyleGAN2 model is proposed~\cite{karras2020analyzing}.
To make the training progress of StyleGAN2 stable with limited data, StyleGAN2-ADA~\cite{karras2020training} introduces an adaptive discriminator augmentation mechanism.
The latent space of these models contains rich decoupled semantic directions, used to manipulate various attributes of fake images, such as facial expression, makeup, object color, image style, and more.

In order to leverage the property for real images, GAN inversion methods are proposed.
GAN inversion methods can typically be divided into optimization-based~\cite{abdal2019image2stylegan, creswell2018inverting, abdal2020image2stylegan++, pan2021exploiting} and encoder-based~\cite{richardson2021encoding, tov2021designing, alaluf2021restyle, xu2021generative, bai2022high, yao2022feature, moon2022interestyle}. 
While optimization-based models exhibit high inversion quality, they require numerous optimization steps for each input image~\cite{kim2021exploiting}, resulting in significant time consumption. 
To overcome this limitation, the encoder-based method emerges as a promising solution, which can operate in real-time~\cite{richardson2021encoding, tov2021designing, alaluf2021restyle, xu2021generative, bai2022high, yao2022feature, moon2022interestyle}.
These methods aim to train encoders to map real-world images into a latent space.
Although GAN inversion has achieved remarkable performances for image-based face editing, this technique has not garnered attention in the field of gait attribute editing.
This paper considers the nature of the latent space rich in semantic information that can also be leveraged to benefit gait attribute editing.
However, unlike RGB-based face images, gait sequences with binary silhouettes possess sparse structural information on the spatial dimension and more individual motion features on the temporal dimension~\cite{ma2022identity}.
This distinction makes the gait inversion particularly challenging.

\begin{figure*}[t]
    \centering
    \includegraphics[width=\linewidth]{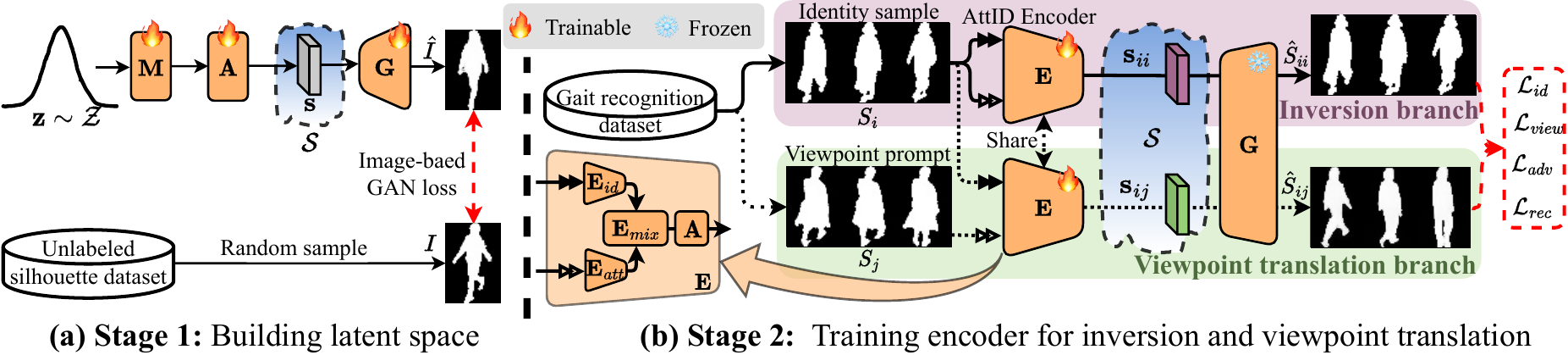}
    \caption{GaitEditor's overall training architecture. The training process involves two stages. (a) An image-based GAN model is trained on an unlabeled silhouette dataset to establish a latent space. (b) GaitEditor involves two branches, inversion and viewpoint translation. The first branch involves projecting the source gait sequence into the established latent space, as well as the other branch focuses on altering the viewpoint from the target sample to that of the source sample. To ensure that all the relevant information in gait sequences is preserved, four losses are designed, and they are identity ($\mathcal{L}_{id}$), viewpoint ($\mathcal{L}_{view}$), video adversarial ($\mathcal{L}_{adv}$), and reconstruction ($\mathcal{L}_{rec}$) losses.} 
    \label{fig:fig2_method_overview}
\end{figure*}

\section{Methodology}\label{sec3}
As shown in Fig.~\ref{fig:fig2_method_overview}, the training scheme of our GaitEditor consists of two stages: the construction of a latent space (Stage 1) and the projection of real gait sequences into this latent space while addressing challenges related to viewpoint translation (Stage 2).
Stage 1 involves constructing an editable latent space enriched with semantic information, which will be expounded upon in Subsection~\ref{sec:latent_space_construction}.
Stage 2 focuses on training a unified encoder to project real gait sequences into this latent space and handle the challenges of viewpoint translation, as illustrated in Subsection~\ref{sec:attribute-identity_encoder}.
Subsequently, Subsection~\ref{sec:training_and_loss} explains the training scheme and well-designed loss functions to train GaitEditor, while Subsection~\ref{sec:gait_attribute_editing} details the process of editing gait attributes using our constructed latent space for input gait sequences.
Finally, we elaborate on the application of GaitEditor on gait recognition and anonymization in Subsection~\ref{sec:method_application}.

\subsection{Latent Space Construction} \label{sec:latent_space_construction}
The ability to edit various gait attributes deeply depends on the semantic separability of the latent space learned. 
The StyleGAN-ADA~\cite{karras2020training}, a powerful style-based GAN model, renowned for its compelling disentangled properties in latent spaces, particularly demonstrated in face attribute editing scenarios~\cite{collins2020editing, shen2020interpreting, wu2021stylespace}.
To apply the capability of these latent spaces in gait silhouette images, we re-train StyleGAN-ADA on OU-MVLP dataset~\cite{takemura2018multi} following official network designs~\cite{karras2020training}, as illustrated in Fig.~\ref{fig:fig2_method_overview} (a).
Formally, given a noisy vector $\mathbf{z}$ in the input latent space $\mathcal{Z}$, which follows a Gaussian distribution, the synthetic image $\hat{I}$ can be generated as follows:
\begin{equation}
\label{equ:stylegan}
\hat{I} = \mathbf{G}(\mathbf{A}(\mathbf{M}(\mathbf{z}))), \quad \mathbf{z} \in \mathcal{Z},
\end{equation}
where $\mathbf{G}$ represents the synthesize network. 
The symbol $\mathbf{M}$ denotes a non-linear mapping network that encodes the noisy vector $\mathbf{z}$ into an intermediate latent vector, $\mathbf{A}$ means a linear layer, projecting the intermediate latent vector into $\mathbf{s}$ within the $\mathcal{S}$ space~\cite{wu2021stylespace}, and $\mathbf{G}$ presents the generator network.
It is worth mentioning that this training process does not rely on identity-level labels.
Given that real silhouette images are typically resized to $64 \times 64$, with binary pixel values, we adopt this size as the default output resolution for the generator $\mathbf{G}$.
Additionally, an unlabeled silhouette image dataset and an image discriminator are employed for adversarial learning. 
After training at this stage, the $\mathcal{S}$ latent spaces are constructed, and generator $\mathbf{G}$ can receive a latent vector $\mathbf{s}$ as input to generate a silhouette image.

\begin{figure}[t]
    \centering
    \includegraphics[width=\linewidth]{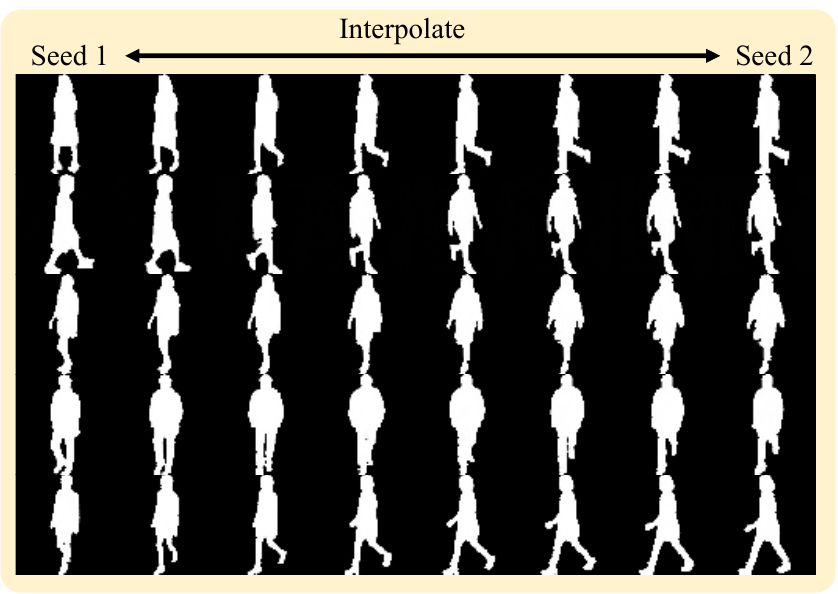}
    \caption{Interpolation results for our build latent space. Each row sequence showcases the stages between two unique random noises. The first and last images in every row are the synthetic visualizations generated from these individual noises.} 
    \label{fig:interpolate}
\end{figure}

To show the efficiency of disentangling gait attributes within our latent space constructed from binary gait silhouettes, we conduct interpolation between two input noises, and use the intermediate results to generate images.
As shown in Fig.~\ref{fig:interpolate}, while performing interpolation between two synthetic silhouette images, the attributes of the generated images have shown a smooth transition, while other attributes remain fixed.
This phenomenon demonstrates that the latent space constructed on binary gait silhouettes also has rich semantic information.

In the face editing, there are many StyleGAN-based editing methods to find semantic directions in their latent spaces, 
such as StyleSpace~\cite{wu2021stylespace}, SeFa~\cite{shen2021closed}, InterFaceGAN~\cite{shen2020interpreting}.
In our study, StyleSpace~\cite{wu2021stylespace} is leveraged to find semantic directions that can edit corresponding image regions of a silhouette image.
Through a detailed manual screening process, we identify ten directions linked to ten specific pedestrian attributes, including gender-like, body size, age-like and clothing types. 
It is worth noting that the connotations of these attributes are inherently subjective, and ground in human perception.
For instance, the attribute of femininity is explained when the pedestrian's hair length is observed to be longer, a trend commonly observed in the OU-MVLP dataset.
The validity and perceptual consistency of these attributes are rigorously validated through a user study. 
The findings are illustrated in Subsection~\ref{sec:user_study}.

Additionally, in the screening process, we discovered that manipulating the latent code in a specific direction only results in a slight change in the filming viewpoint.
The primary reason is variations in the viewpoint significantly affect the appearance of the silhouette, \textit{i.e.} only manipulating a single direction cannot handle the viewpoint translation task.
Manipulating synthetic silhouettes is not our ultimate objective, and our focus is on editing real gait sequences within the latent space.
In accordance with the aforementioned insights, we introduce a dual-stream encoder architecture termed the Attribute-Identity (AttID) Encoder $\mathbf{E}$ to project a real gait sequence into established latent space and handle the challenge of viewpoint translation.

\subsection{Attribute-Identity Encoder} \label{sec:attribute-identity_encoder}
\begin{figure*}[t]
    \centering
    \includegraphics[width=0.7\linewidth]{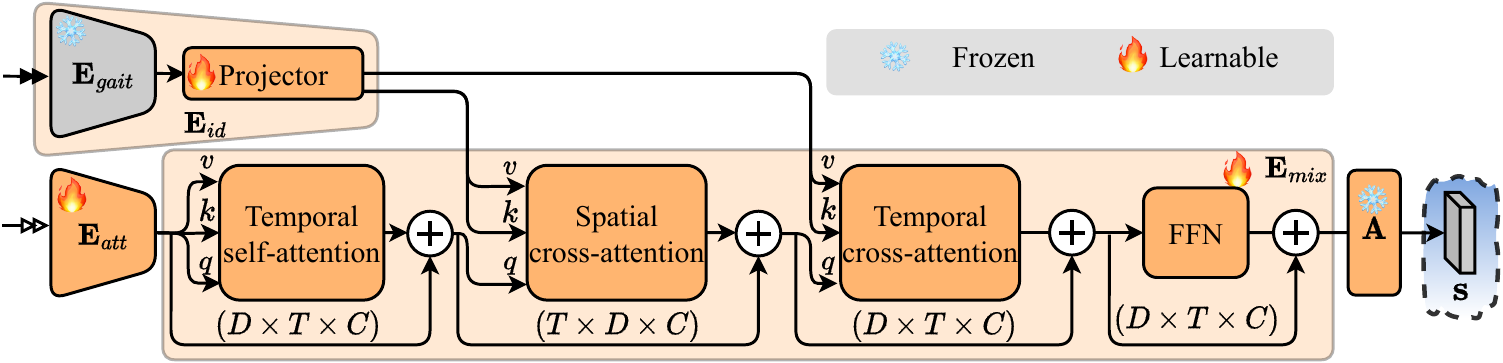}
    \caption{The architecture of Attribute-Identity encoder.} 
    \label{fig:fig3_attidencoder}
\end{figure*}

In Stage 2, GaitEditor encompasses two branches: the inversion branch (purple background) and the viewpoint translation branch (green background), as shown in Fig.~\ref{fig:fig2_method_overview} (b). 
The former is for projecting real gait sequences into constructed latent spaces, and the latter is for handling the viewpoint translation challenge.
In this stage, both pre-trained liner layer $\mathbf{A}$ and generator $\mathbf{G}$ will be frozen all the time.
For a specialized gait recognition dataset, two different real gait sequences $S_{i}$ and $S_{j}$ are randomly sampled, denoted as identity sample and viewpoint prompt, respectively. 
For the inversion task, only the $S_{i}$ is input into the AttID Encoder $\mathbf{E}$, then input into liner layer $\mathbf{A}$ to obtain latent codes $\mathbf{s}_{ii}$.
Simultaneously, for the viewpoint translation task, AttID Encoder $\mathbf{E}$ takes both the $S_i$ and $S_j$  as input and then the results of $\mathbf{E}$ input into liner layer $\mathbf{A}$ to generate the viewpoint-swapped latent code $\mathbf{s}_{ij}$. 
Specifically, AttID Encoder $\mathbf{E}$ comprises three meticulously designed blocks: identity extractor $\mathbf{E}_{id}$, attribute extractor $\mathbf{E}_{att}$, and feature mix block $\mathbf{E}_{mix}$, which responsible for extracting identity, attribute information, and fusing them, respectively.
Formally, the process of obtaining intermediate latent vectors is expressed as follows:
\begin{equation}
\begin{aligned}
    \mathbf{s}_{ii} &= \mathbf{A}(\mathbf{E}(S_{i}, S_{i})) = \mathbf{A}(\mathbf{E}_{mix}(\mathbf{E}_{id}(S_i), \mathbf{E}_{att}(S_i))), \\
    \mathbf{s}_{ij} &= \mathbf{A}(\mathbf{E}(S_{i}, S_{j})) = \mathbf{A}(\mathbf{E}_{mix}(\mathbf{E}_{id}(S_i), \mathbf{E}_{att}(S_j))) .
\end{aligned}
\end{equation}
Finally, the frozen $\mathbf{G}$ then decodes $\mathbf{s}_{ii}$ and $\mathbf{s}_{ij}$ into the fake gait sequences $\hat{S}_{ii}$, and $\hat{S}_{ij}$ frame-by-frame.
Formally, the process of projection and generation is expressed as follows:
\begin{equation}
\begin{aligned}
    \hat{S}_{ii} &= \mathbf{G}(\mathbf{s}_{ii}), \\
    \hat{S}_{ij} &= \mathbf{G}(\mathbf{s}_{ij}).
\end{aligned}
\end{equation}
It is noteworthy that $\hat{S}_{ij}$ contains a synthesized viewpoint corresponding to the viewpoint prompt $S_{j}$ while retaining the identity information of the identity sample $S_{i}$.

In the aspect of model architectures, for $\mathbf{E}_{id}$, to reduce the intra-class divergences of gait distribution, we utilize a pre-trained appearance-based gait recognition model as the backbone ($\mathbf{E}_{gait}$), and freeze it when GaitEditor training.
As such, to accomplish the gait sequence reconstruction task in Stage (ii), the learnable  $\mathbf{E}_{att}$ will be forced to act as a supplement of $\mathbf{E}_{id}$, \textit{i.e.} try its best to exploit the diversity of input gait sequence. 
Meanwhile, for $\mathbf{E}_{att}$, following~\cite{richardson2021encoding}, we use a pSp-like network taking the ResNet18~\cite{he2016deep} as the backbone.
Practically, $\mathbf{E}_{att}$ processes each silhouette frame by frame from the input viewpoint prompt, yielding features with dimensions $T \times D \times C$, where $T$, $D$, and $C$ respectively denote sequence length, embedding dimension, and channel dimension.
Meanwhile, $\mathbf{E}_{gait}$ compresses the input identity sample into identity embedding frame by frame with the shape of $ T \times P_{gait} \times C_{gait}$, where $P_{gait}$, $C_{gait}$ respectively represent the dimension of part and channel~\cite{fan2023opengait}.
To align the dimensions of the identity and attribute features, we introduce a learnable projector module, which is a non-linear module. 
Through this projector module, the identity feature attains the shape of $T \times D \times C$.

In the architecture illustrated in Fig.~\ref{fig:fig3_attidencoder}, our $\mathbf{E}_{mix}$ comprises four key components: temporal self-attention, spatial cross-attention, temporal cross-attention, and feed-forward blocks.
These components synergistically work to mix identity and attribute embeddings. 
Firstly, the output from $\mathbf{E}_{att}$ is fed into the temporal self-attention module to capture the inter-frame relationships, thereby enhancing the walking faithfulness and inter-frame consistency of the synthetic gait sequences. 
Subsequently, the identity embedding, extracted by $E_{id}$, serves as both the key ($k$) and value ($v$), while the result of the addition between temporal self-attention and attribute embedding is utilized as query ($q$).
These elements are then input into spatial and temporal cross-attention modules, respectively. 
This process facilitates the intra- and inter-frame fusion of the identity embedding with the embedding extracted from the viewpoint prompt.
In this framework, the viewpoint prompt provides coarser-grained global information, while the identity sample offers finer-grained local details. 
By employing the cross-attention modules, we can effectively mix information of different granularities.

\subsection{Loss Functions and Training Scheme} \label{sec:training_and_loss}
In our training stage, four loss functions are conducted,  \textit{i.e.}, the reconstruction, identity, video adversarial, and viewpoint classification loss.

\textbf{Reconstruction loss. } $\mathcal{L}_{rec}$ can directly preserve the visual authenticity of the generated sequences. 
Therefore, we introduce the pixel-level and perception-level reconstruction terms, which are formulated by
\begin{equation}
\begin{aligned}
    \mathcal{L}_{rec} &= \mathcal{L}_{pix} + \mathcal{L}_{per}, \\
    \mathcal{L}_{pix} &= \left\|\hat{S}_{ii} - S_{i} \right\|_2, \\
    \mathcal{L}_{per} &= \left\|V(\hat{S}_{ii}) - V(S_{i}) \right\|_2, 
\end{aligned}
\end{equation}
where $\left\|\cdot \right\|_2$ and $V$ respectively denote the $L_2$ distance function and VGG-16 model trained on ImageNet~\cite{deng2009imagenet}. 
$\mathcal{L}_{per}$ presents the LPIPS loss~\cite{zhang2018perceptual} widely used to maintain the similarity in visual attributes.
Note that the reconstruction process works only in the case of two input gait sequences being identical, \textit{i.e.}, inversion branch case. 

\textbf{Video adversarial loss.} To ensure the faithfulness of the synthesized gait sequences, we develop a video-based adversarial loss, which utilizes the least squares GAN loss~\cite{mao2017least} and can be formulated as:
\begin{equation}
\begin{aligned}
    \mathcal{L}_{adv}^{\mathbf{D}_{vid}} &= \frac{1}{2} \sum\limits_{y \in \{i,j\}}(\mathbb{E}_{S_{i}\sim \mathcal{D}}\left[(\mathbf{D}_{vid}(S_{i}) - 1)^2 \right] \\
    &+ \mathbb{E}_{\hat{S}_{iy}\sim \hat{\mathcal{D}}}\left[(\mathbf{D}_{vid}(\hat{S}_{iy}))^2 \right]), \\
    \mathcal{L}_{adv}^\mathbf{E} &= \frac{1}{2}\sum\limits_{y \in \{i,j\}}\mathbb{E}_{\hat{S}_{iy}\sim \hat{\mathcal{D}}}\left[(\mathbf{D}_{vid}(\hat{S}_{iy})-1)^2 \right],   \\
\end{aligned}
\label{eq:dis_vid}
\end{equation}
where $\mathbf{D}_{vid}$ represents the video discriminator, which is a patch discriminator with 3D convolution blocks~\cite{tulyakov2018mocogan}.
$\mathcal{D}$ and $\hat{\mathcal{D}}$ respectively mean the real and synthetic gait dataset. 
During the adversarial learning process, $\mathcal{L}_{adv}^{\mathbf{D}_{vid}}$ and $\mathcal{L}_{adv}^\mathbf{E}$ are optimized alternatively with respectively updating the training parameters of video discriminator $\mathbf{D}_{vid}$ and AttID encoder $\mathbf{E}$. 
The adversarial process works in all branches, \textit{i.e.} inversion and viewpoint translation branches.

\textbf{Identity loss.} 
To preserve the identity consistency for the synthetic gait sequences, we leverage an off-shelf gait recognition model $\mathbf{E}_{gait}$ again, which has been mentioned in Fig.~\ref{fig:fig3_attidencoder}, to constitute the identity constraint.
Specifically, for the inversion branch, we leverage MSE loss and cosine similarity loss to close $S_{i}$ and $\hat{S}_{ii}$ in the identity embedding space.
\begin{equation}
\begin{aligned}
    \mathcal{L}_{id}^{inv} &= (1 - \left\langle \mathbf{E}_{gait}(\hat{S}_{ii}), \mathbf{E}_{gait}(S_{i}) \right\rangle) \\ 
    &+ \left\|\mathbf{E}_{gait}(\hat{S}_{ii}) - \mathbf{E}_{gait}(S_{i}) \right\|_2 , 
\end{aligned}
\label{eq:id_loss}
\end{equation}
where $\left\langle \cdot, \cdot \right\rangle$ represents the cosine similarity function.

For the branch of viewpoint translation, we employ an InfoNCE loss~\cite{gutmann2010noise} within the identity embedding space. 
Specifically, considering a minibatch with a batch size $N$, and defining $(S_{ij}^{n}, S_{i}^{n})$ as the positive pair, and both $(S_{ij}^{n}, S_{i}^{m})$ with $n \neq m$ and $(S_{ij}^{n}, S_{j}^{l})$ as the negative pairs, where $n, m, l \in \{ 1,2, \cdots, N \}$, we formulate the identity InfoNCE loss as follows:
\begin{equation}
\begin{aligned}
    &\mathcal{L}_{id}^{vt} = -\frac{1}{N}\sum \limits_{n=1}^{N} [\log(M_{id}^{n})], \\
    &M_{id}^{n} = \frac{Q(\hat{S}_{ij}^{n}, S_{i}^{n})}
                          {\sum \limits_{m=1}^{N} Q(\hat{S}_{ij}^{n}, S_{i}^{m}) + \sum \limits_{l=1}^{N}Q(\hat{S}_{ij}^{n}, S_{j}^{l})},  \\
    &Q(A, B) = \exp(\mathrm{sim}(\mathbf{E}_{gait}(A), \mathbf{E}_{gait}(B))/\tau),
\end{aligned}
\end{equation}
where $$\mathrm{sim}(\mathbf{E}_{gait}(A), \mathbf{E}_{gait}(B)) = \frac{\mathbf{E}_{gait}(A)^\top \cdot \mathbf{E}_{gait}(B)}{\left\|\mathbf{E}_{gait}(A)\right\|_2 \left\|\mathbf{E}_{gait}(B) \right\|_2}$$ is cosine similarity, and $\tau > 0$ denotes the temperature hyperparameter set to 16.

The total identity loss is the sum of these two items, \textit{i.e.}:
\begin{equation}
    \mathcal{L}_{id} = \mathcal{L}_{id}^{inv} + \mathcal{L}_{id}^{vt}.
\end{equation}

\textbf{Viewpoint loss.} 
To maintain the viewpoint consistency between the generated gait sequence and the viewpoint prompt, the attribute loss should be introduced.
Specifically, we use a viewpoint classifier $\mathbf{E}_{view}$ trained on OU-MVLP~\cite{takemura2018multi} to build the loss:
\begin{equation}
\begin{aligned}
   \mathcal{L}_{view} &= D_{KL}(\mathbf{E}_{view}(\hat{S}_{ii})||\mathbf{E}_{view}(S_i)) \\
                      &+ D_{KL}(\mathbf{E}_{view}(\hat{S}_{ij})||\mathbf{E}_{view}(S_j)), 
\end{aligned}
\label{eq:view}
\end{equation}
where $D_{KL}(\cdot||\cdot)$ denotes the Kullback-Leibler divergence.

\textbf{Training scheme and total loss.}
In our study, we first train the inversion branch and then train both the inversion and viewpoint translation branches.
The total loss is defined as a weighted combination of different losses:
\begin{equation}
\begin{aligned}
   \mathcal{L}_{total} &= \lambda_{rec}\mathcal{L}_{rec} + \lambda_{id}\mathcal{L}_{id} + \lambda_{view}\mathcal{L}_{view} \\
   &+ \lambda_{adv}(\mathcal{L}_{adv}^{\mathbf{D}_{vid}} + \mathcal{L}_{adv}^{\mathbf{E}}).
\end{aligned}
\label{eq:total}
\end{equation}
We set $\lambda_{rec}=1$, $\lambda_{id}=0.6$, $\lambda_{view}=0.1$, $\lambda_{adv}=0.4$ for only the inversion branch training, and $\lambda_{rec}=0.6$, $\lambda_{id}=1.0$, $\lambda_{view}=1.2$, $\lambda_{adv}=0.1$ for both inversion and viewpoint translation branches training.

\begin{figure}[t]
    \centering
    \includegraphics[width=0.9\linewidth]{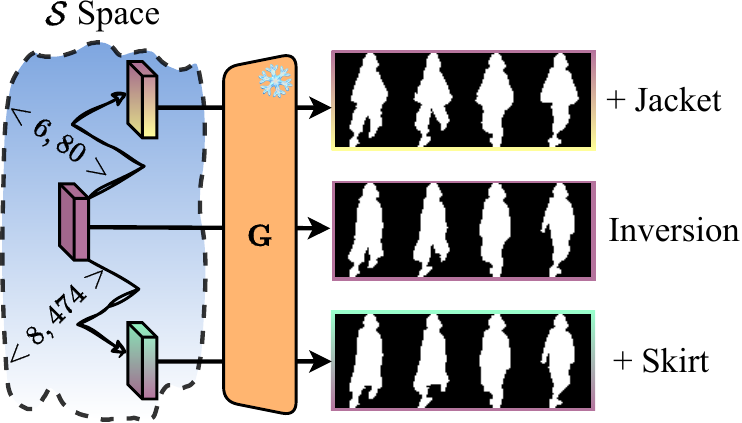}
    \caption{The schematic of editing non-viewpoint attributes.} 
    \label{fig:fig5_s_editing}
\end{figure}

\begin{figure*}[t]
    \centering
    \includegraphics[width=0.9\linewidth]{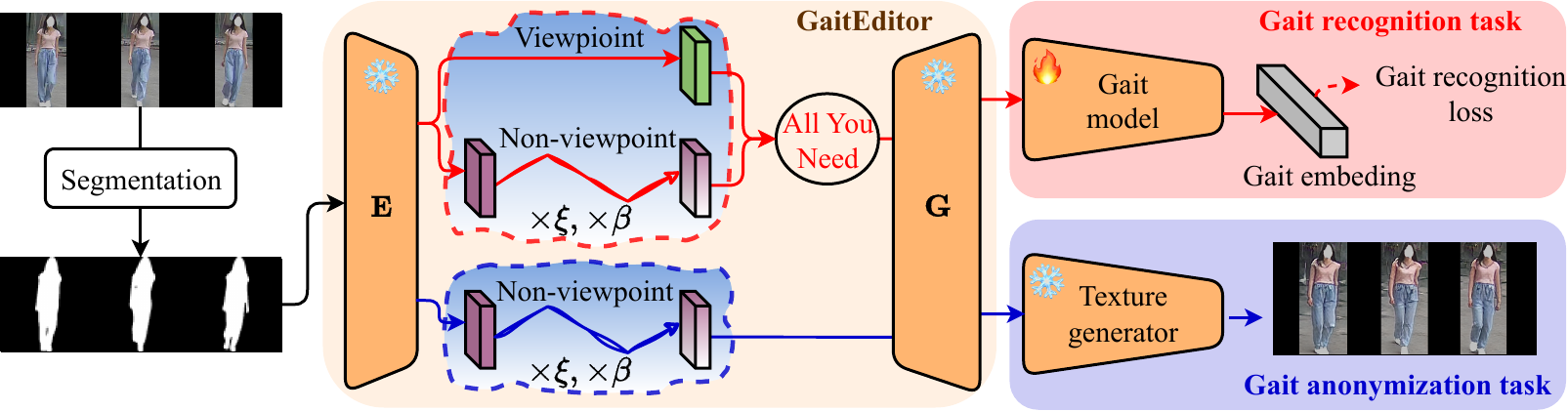}
    \caption{The overview of GaitEditor's applications. Where $\xi$ and $\beta$ denote the editing frequency and strength, respectively.} 
    \label{fig:fig4_applications}
\end{figure*}

\begin{figure*}[t]
    \centering
    \begin{subfigure}{.32\textwidth}
        \centering
        \includegraphics[width=\linewidth]{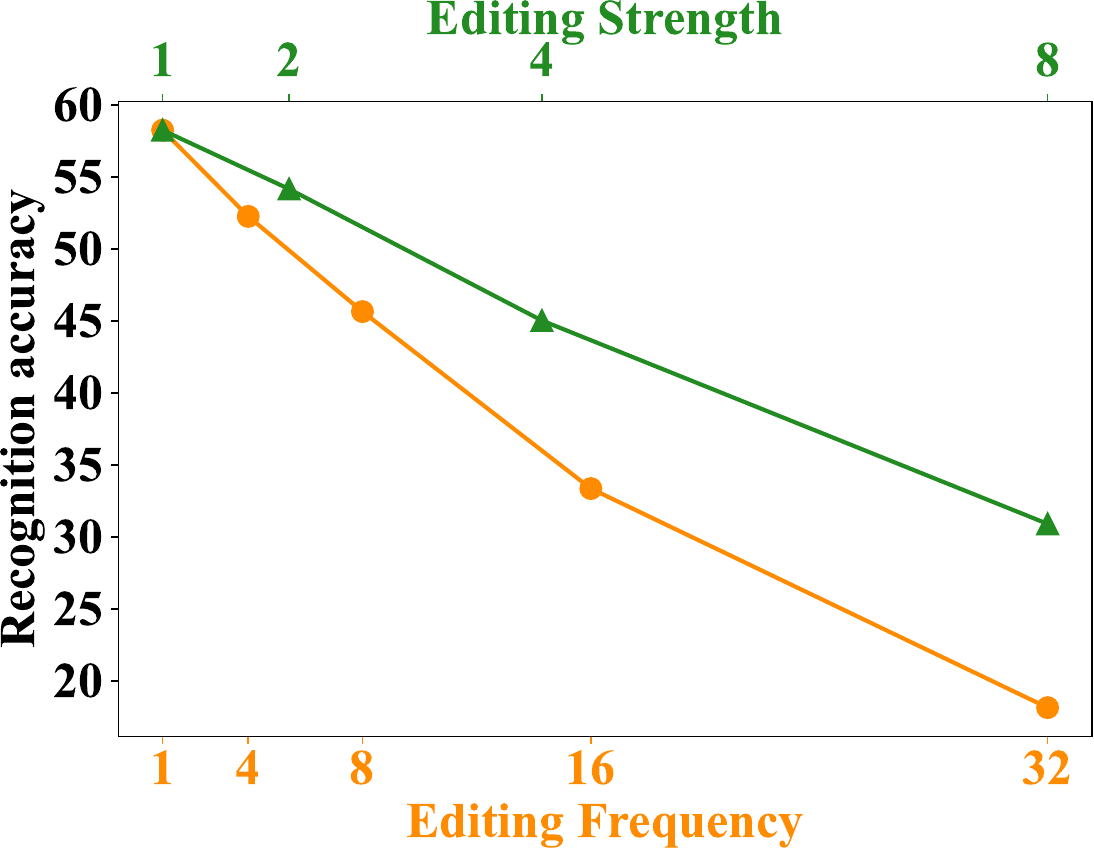}
        \caption*{(a) Recognition accuracy metric.}
        \label{fig:rank_1}
    \end{subfigure}%
    \hfill
    \begin{subfigure}{.32\textwidth}
        \centering
        \includegraphics[width=\linewidth]{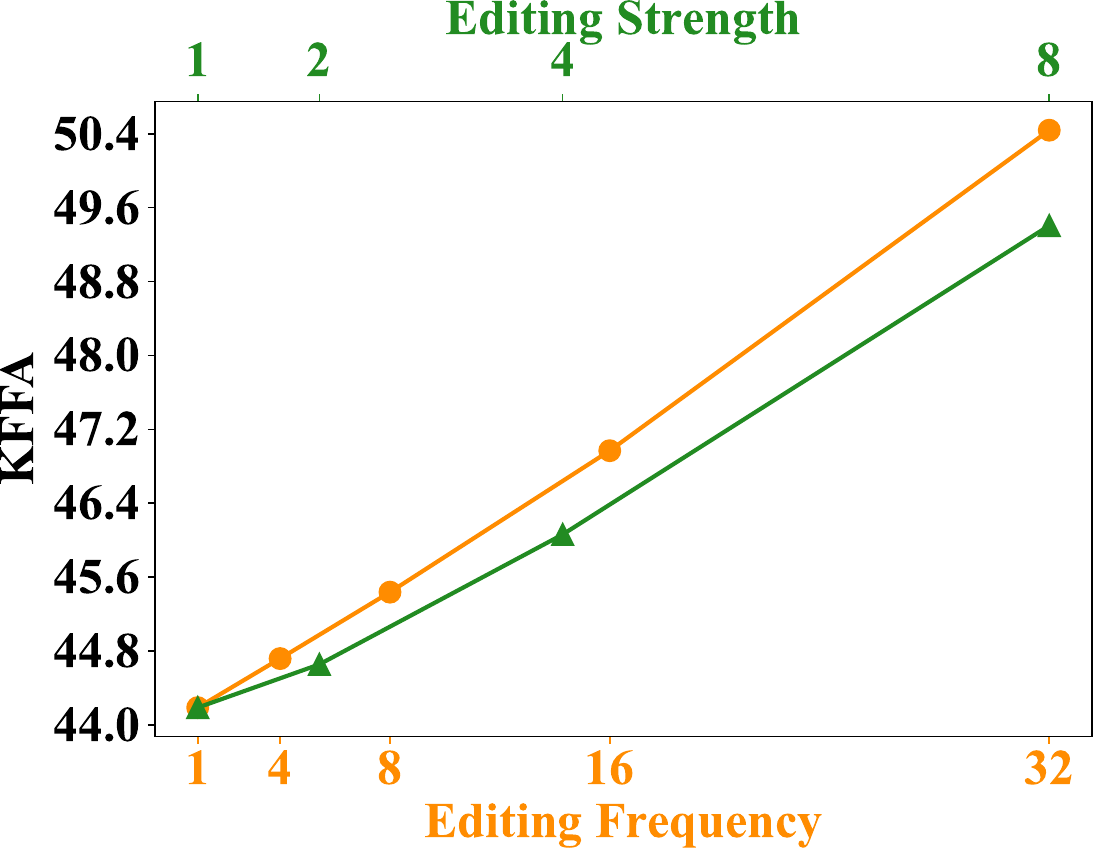}
        \caption*{(b) KFFA metric.}
        \label{fig:kffa}
    \end{subfigure}%
    \hfill
    \begin{subfigure}{.32\textwidth}
        \centering
        \includegraphics[width=\linewidth]{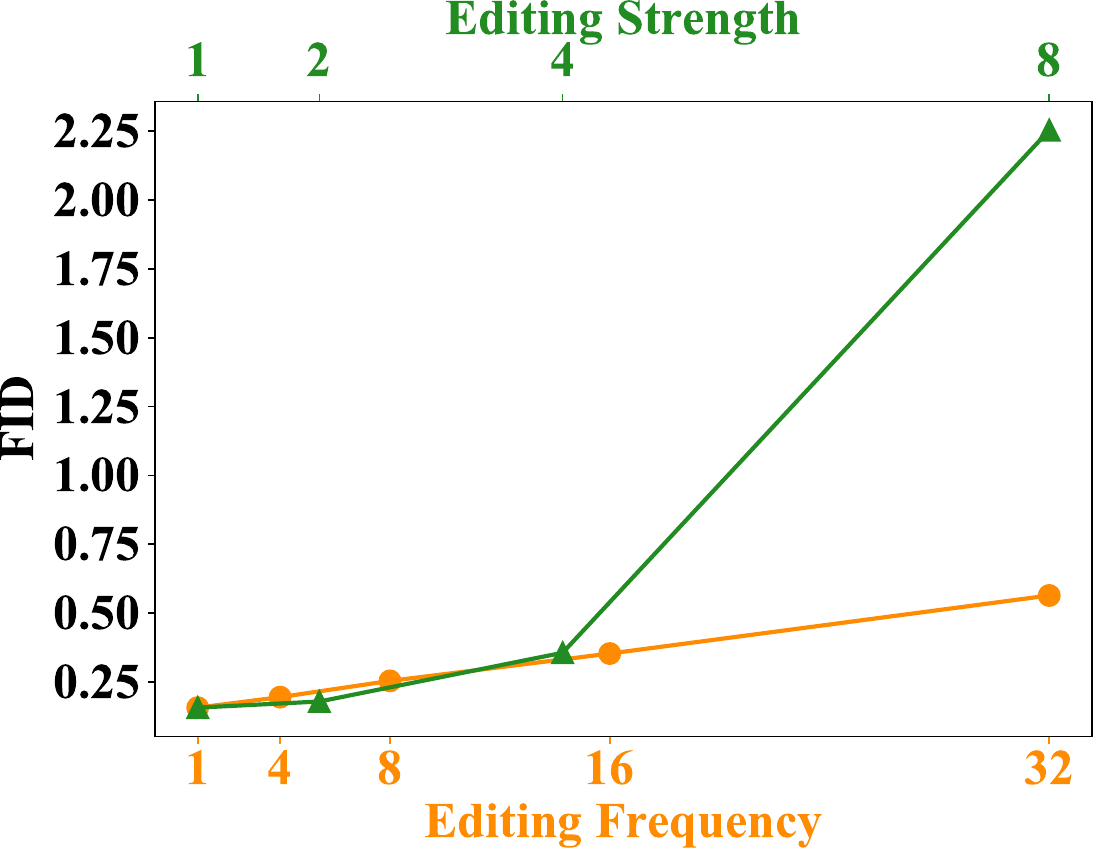}
        \caption*{(c) FID metric.}
        \label{fig:sil_fid}
    \end{subfigure}
    \caption{The performance on three metrics, (a) Recognition (rank-1) accuracy, (b) KFFA, and (c) FID, under varying editing strengths and frequencies.}
    \label{fig:fig6_accuracy}
\end{figure*}

\subsection{Gait Attribute Editing} \label{sec:gait_attribute_editing}
Through the aforementioned training process, GaitEditor demonstrates the capability to effectively project a real gait sequence into the constructed $\mathcal{S}$ space and handle the challenge of viewpoint attribute editing.

\textbf{Editing non-viewpoint attributes.} 
In this progress, we want to use ten directions that have been selected in Stage 1 for editing real gait sequences.
Therefore, the pipeline of the inversion branch is used.
This progress will receive two identical real gait sequences as input to project it into $\mathcal{S}$ space by $\mathbf{E}$, and then we manipulate the semantic directions in $\mathcal{S}$ space to edit non-viewpoint attributes.
A semantic direction in $\mathcal{S}$ space typically can be denoted as $<l, c>$, where $l$ and $c$ represent the layer and channel dimension, respectively.
For a latent code $s$, we can manipulate their $l$-th layer and $c$-th channel to edit corresponding gait attributes.
Fig.~\ref{fig:fig5_s_editing} illustrates the manipulation of the latent code $s_{ii}$ using a specific direction $<l, c>$.
The progress of manipulation can be formulated as:
\begin{equation}
    s_{ii}[l][c] \mathrel{+}= \sigma(s_{ii}[l])[c] * (\alpha * \beta),
    \label{eq:s_editing}
\end{equation}
where $\sigma(s_{ii}[l])$ means the standard deviation of each channel for $l$-th layer, $\alpha$ indicates the number of editing strength, $\beta$ represents the coefficient of editing strength.
In our case, the direction of $< 8, 474 >$ and $< 6, 80 >$ indicates the manipulation of wearing a shirt and a jacket, respectively.

\textbf{Editing viewpoint attributes.} 
In this progress, the pipeline of the viewpoint translation branch is used, and the $\mathbf{E}$ will receive two different real gait sequences, $S_i$ and $S_j$, as input.
Then, $S_i$ is viewed as the identity sample, which mainly provides identity information, and $S_j$ is viewed as the viewpoint prompt, which mainly provides viewpoint information.
Finally, though our $\mathbf{E}$ and $\mathbf{G}$, the fake gait sequence $\hat{S}_{ij}$ is generated, with the viewpoint of $S_j$ and the identity of $S_i$. 

\subsection{Applications} \label{sec:method_application}

As depicted in Fig.~\ref{fig:fig4_applications}, GaitEditor can act as a plug-and-play module that can be readily employed in the workflow of gait recognition and anonymization tasks. 
In our experiments, we will leverage the full spectrum of semantic directions within the $\mathcal{S}$ space. 
Although Stage 1 identifies ten semantic directions capable of editing ten distinct gait attributes, the $\mathcal{S}$ space encompasses a broader range of other semantic directions.
These vectors, to varying extents, edit specific regions in silhouette images, thereby providing extensive diversity for further exploration in gait recognition and anonymization tasks.

In order to further leverage these rich semantic directions, a strategy of randomized editing iterations is introduced.
Specifically, we perform multiple edits on a gait sequence, each time randomly selecting a semantic direction and adjusting different strength coefficients $\beta$, as outlined in Equation~\ref{eq:s_editing}.
For clarity, we name the editing times as editing frequency, $\xi$,  and the strength coefficients as editing strength, $\beta$.
Our experimental observations reveal that both the editing strength and frequency substantially influence the diversity in factors and the degree of identity preservation within the generated gait sequences, as shown in Fig.~\ref{fig:fig6_accuracy}. 
In order to quantitatively validate this point, we use three metrics on CCPG dataset~\cite{li2023depth}: rank-1 accuracy, Known Factor Feature Angle (KFFA)~\cite{li2022contrastive}, and Fr\'{e}chet Inception Distance (FID)~\cite{heusel2017gans}, at various editing frequencies and strengths, thereby respectively evaluating the degree of identity preservation, diversity in factors and quality of generated gait sequences.

Specifically, for evaluating rank-1 accuracy, GaitEditor is utilized to edit the entire test probe set of CCPG dataset~\cite{li2023depth} with different conditions, thereby evaluating the recognition accuracy of the GaitBase model~\cite{fan2023opengait}. 
For KFFA, GaitSSB~\cite{fan2023learning} is utilized to transfer the generated CCPG dataset samples into their latent space.
Similarly, a 192-channel variant of InceptionV3~\cite{szegedy2016rethinking} is employed for FID computation.
As shown in Fig.~\ref{fig:fig6_accuracy} (a), the results reveal a negative correlation between recognition accuracy and editing degrees, indicating that increased editing frequency or strength decreases accuracy due to the introduced variability. 
Conversely, as shown in Fig.~\ref{fig:fig6_accuracy} (b), the KFFA metric displays a positive trend, suggesting that higher editing frequencies or strengths enhance the diversity of gait features. 
The FID score increases with greater editing but remains within acceptable limits for frequencies below 32 and strengths below 4, as shown in Fig.~\ref{fig:fig6_accuracy} (c). 
Accordingly, we meticulously control the editing strengths and frequencies to respectively improve performance in gait recognition and anonymization tasks.

\subsubsection{Gait Recognition}
In this application scenario, GaitEditor serves as an online data augmentation tool during the training phase of gait models, aiming to improve recognition accuracy across various challenge benchmarks.
A significant challenge in gait recognition is the limited diversity of factors in training sets, detrimentally impacting recognition performance.
GaitEditor tackles this issue by modulating the editing frequency and strength, as depicted in Fig.\ref{fig:fig5_s_editing} (upper branch).
For each gait sequence, editing is conducted multiple times, involving random modifications of both viewpoint and non-viewpoint attributes with an editing strength coefficient $\beta$.
Regarding non-viewpoint attribute editing, GaitEditor first projects a real or edited gait sequence into the $\mathcal{S}$ space, resulting in a latent code.
This code is then manipulated multiple times in different directions, generating the final gait sequence.
For viewpoint attributes, GaitEditor uses the AttID encoder to adaptively transfer the viewpoint from a prompt to the identity sample. 
To maintain identity consistency while introducing factor diversity, we set $\xi$ ranging from 1 to 8 and $\beta$ from 0 to 2. The specifics of these parameter settings are contingent upon the training dataset, as depicted in Table~\ref{tab:finetuning_setting}.

\begin{table*}[t]
\centering
\caption{The results of identity consistency achieved through editing various attributes are presented. The term "ID-Similarity" refers to the Cosine Similarity between edited and real gait sequences. The '\textcolor[rgb]{0.80,0.70,0.00}{Inversion}' involves reconstructing the input sample using our GaitEditor, representing the upper bound of identity similarity. Conversely, '\textcolor[rgb]{0.00,0.50,0.00}{Negative-pairs}' denotes the computation of identity similarity between two distinct individuals, serving as the lower bound of identity similarity.}
\label{tab:identity_consistency}
\resizebox{0.85\linewidth}{!}{%
\centering
\begin{tabular}{@{}c|cccccc@{}}
\toprule
Editing Attributes                                                        & \textcolor[rgb]{0.80,0.70,0.00}{Inversion}                                          & Fatness         & Thinness        & Feminine        & Older           & Younger                                                                           \\ \midrule
\begin{tabular}[c]{@{}c@{}}ID-Similarity \\ (CCPG $|$ OU-MVLP)\end{tabular} & \textcolor[rgb]{0.80,0.70,0.00}{0.9785} $|$ \textcolor[rgb]{0.80,0.70,0.00}{0.9336} & 0.9399 $|$ 0.7985 & 0.9327 $|$ 0.8026 & 0.9322 $|$ 0.8319 & 0.9457 $|$ 0.8500 & 0.9329 $|$ 0.8610                                                                   \\ \midrule
Editing Attributes                                                        & Jacket                                                                              & Skirts          & Pants           & Hoodie          & Viewpoints      & \textcolor[rgb]{0.00,0.50,0.00}{Negative-pairs}                                   \\ \midrule
\begin{tabular}[c]{@{}c@{}}ID-Similarity \\ (CCPG $|$ OU-MVLP)\end{tabular} & 0.9576 $|$ 0.8039                                                                     & 0.9513 $|$ 0.7872  & 0.9295 $|$ 0.7612 & 0.9202 $|$ 0.7167 & 0.9380 $|$ 0.7813 & \textcolor[rgb]{0.00,0.50,0.00}{0.8806} $|$ \textcolor[rgb]{0.00,0.50,0.00}{0.5153} \\ \bottomrule
\end{tabular}%
}
\end{table*}

\begin{table*}[t]
\centering
\caption{The attributes correctness for different attributes. We conduct a user study to evaluate. \textit{w/} Exp. means the subjects have gait experience and \textit{w/o} Exp. means the subjects have no gait experience.}
\label{tab:user_study}
\resizebox{0.85\linewidth}{!}{%
\centering
\begin{tabular}{@{}c|ccccccccccc@{}}
\toprule
\multirow{2}{*}{} & \multicolumn{11}{c}{Accuracy (\%)}                                                                                         \\ \cmidrule(l){2-12} 
                  & Fatness & Thinness & Masculine & Feminine & Older & Younger & Jacket & Skirt & Pants & \multicolumn{1}{c|}{Hoodie} & Mean  \\ \midrule
\textit{w/} Exp.  & 72.41   & 88.37    & 74.42     & 82.61    & 69.35 & 76.47   & 90.57  & 94.55 & 88.24 & \multicolumn{1}{c|}{77.59}  & 81.46 \\
\textit{w/o} Exp. & 73.33   & 82.69    & 45.00     & 68.52    & 58.33 & 55.74   & 75.00  & 85.42 & 86.27 & \multicolumn{1}{c|}{82.00}  & 71.23 \\ \midrule
Mean              & 72.87   & 85.53    & 59.71     & 75.57    & 63.84 & 66.11   & 82.79  & 89.99 & 87.26 & \multicolumn{1}{c|}{79.80}  & 76.34 \\ \bottomrule
\end{tabular}%
}
\end{table*}

\subsubsection{Gait Anonymization}
In this specific application scenario, GaitEditor is primarily designed to anonymize identities within gait sequences while ensuring minimal visual distortion.
According to the results of Fig.~\ref{fig:fig6_accuracy}, we observe a decrease in recognition accuracy corresponding to an increased $\xi$ or $\beta$, but have a negligible effect on image quality. 
Leveraging this insight, GaitEditor enables anonymising identity within gait sequences without significantly modifying the individual's appearance.
Specifically, the process begins with the detection and segmentation of an input gait sequence to extract its silhouette. 
GaitEditor then edits this silhouette sequence's attributes in the $\mathcal{S}$ space multiple times with an editing strength, $\beta$.
In our study, we set $\xi$ to 32 and $\beta$ to 1 for the gait anonymization task. 
Subsequently, a pre-trained image-to-image generative network, VQGAN~\cite{esser2021taming}, transforms the edited silhouette sequence into a realistic RGB video format. 
This workflow is detailed in Fig.~\ref{fig:fig4_applications} (lower branch).

\section{Experiments}\label{sec4}

\subsection{Datasets and Evaluation Metrics}
In our study, we employ four public gait datasets involving the most widely-used dataset OU-MVLP~\cite{takemura2018multi}, the largest cloth-changing dataset CCPG~\cite{li2023depth}, the most factor-complete dataset CASIA-E~\cite{song2022casia}, and the most up-to-date popular in the wild dataset Gait3D~\cite{zheng2022gait}. 
Unless otherwise stated, our implementation follows official protocols, including the training/testing and gallery/probe set partition strategies.

\textbf{OU-MVLP}~\cite{takemura2018multi} stands out as one of the most extensive indoor gait datasets available.  
It encompasses data from 10,307 individuals, with each participant's data reflecting a singular walking scenario, specifically normal walking.
The dataset captures each walking instance through videos shot from 14 positioned angles, ranging between [$0^\circ$, $90^\circ$] and [$180^\circ$, $270^\circ$]. 
Therefore, there are 28 sequences per subject.
Following the dataset's official protocol, we have divided the subjects into two groups: 5,153 for training and 5,154 for testing. 
During the testing phase, the first sequence (NM\#0) of each subject is used for the probe, while the subsequent sequence (NM\#1) is utilized for the gallery.

\textbf{CCPG}~\cite{li2023depth} is the largest cloth-changing gait dataset. 
This dataset comprises 200 subjects, each subject wearing seven different clothing styles, such as complete outfit alterations, top changes, pant changes, and scenarios involving carrying bags, and each subject contains 10 different camera viewpoints.
The dataset, comprising over 16,000 sequences, captures both RGB and silhouette modalities.
These subjects and sequences are categorized into two distinct groups for training and testing purposes, with the first 100 subjects allocated for training and the remaining 100 for testing. 
In the test subset, we have four cloth-changing settings following official guidelines.
1) Cloth-changing (\textbf{CL}): For probing, we use sequences U0D0 and U0D0BG, whereas U1D1, U2D2, and U3D3 sequences are designated for the gallery set. 
U0D0BG indicates different clothing combinations, which "U" means the upper-clohting, "D" means the pant, "BG" means bag and "0" means the clothing number.
2) Ups-changing (\textbf{UP}): The probe set involves sequences of U3D3, contrasted with U0D3 sequences for the gallery.
3) Pants-changing (\textbf{DN}): In this setting, U1D0 sequences are employed for probing, while the gallery is represented by U1D1 sequences.
4) Bag-Changing (\textbf{BG}): We utilize sequences U0D0BG for the probe and U0D0 sequences for the gallery.



\textbf{Gait3D}~\cite{zheng2022gait} was collected in a large supermarket and contains 1,090 hours of high-resolution videos, each with a resolution of 1920×1080 pixels and a frame rate of 25 FPS. 
This dataset is comprised of recordings from 4,000 subjects, containing over 25,000 walking sequences. 
These subjects are categorized into two distinct groups for training and testing purposes, with 3,000 subjects allocated for training and the remaining 1,000 for testing. 
In the test subset, only one sequence from each ID will be selected to build the probe set with 1,000 sequences, while the remaining sequences form a gallery set totalling 5,369 sequences.

For GaitEditor evaluation, we employ Peak Signal-to-Noise Ratio (PSNR), Structural Similarity (SSIM)~\cite{wang2004image}, FID~\cite{heusel2017gans}, Learned Perceptual Image Patch Similarity (LPIPS)~\cite{zhang2018unreasonable}, and Mean Square Error (MSE).
The first two and MSE are used to evaluate the reconstruction performance, FID to evaluate silhouette image quality, and LPIPS to evaluate silhouette image perceptual quality.
For gait recognition evaluation, Rank-1 accuracy is used as the primary evaluation metric.
For gait anonymization evaluation, Rank-1 accuracy and FID are used.

\subsection{Implement Details}
In Stage 1, GaitEditor is trained on the OU-MVLP training set without requiring identity-level annotations. 
The StyleGAN2-ADA~\cite{karras2020training} model \footnote{Publicly available at \url{https://github.com/NVlabs/stylegan2-ada-pytorch.git}} is trained with 16500 ($\sim$3d) tricks. 
We apply various ADA options, such as horizontal flipping, image-space filtering, image corruption, isotropic scaling, integer translation, anisotropic scaling, fractional translation, and rotation of plus or minus ten degrees. 
Other hyper-parameters are set to the same values as in the official protocol~\cite{karras2020training}.
The resulting FID scores~\cite{heusel2017gans}, using a 2048-channel version of InceptionV3, are 2.93, demonstrating the verisimilitude of our generated silhouette images. 

In Stage 2, we freeze the pre-trained StyleGAN2-ADA model and train the AttID Encoder $\mathbf{E}$ and video discriminator $\mathbf{D}_{vid}$. 
Regarding different datasets, we leverage the training set of corresponding datasets to train different Stage 2, respectively.
We use the officially provided GaitBase network~\cite{fan2023opengait} as $\textbf{E}_{id}$, which are pre-trained on corresponding datasets.
We first train the inversion branch at the first $N_{epoch}$ epoch and then train both the inversion and viewpoint translation branches.
For the OU-MVLP dataset, we set $N_{epoch} = 8$, and the total training epoch is 50 ($\sim$25h).
For the CCPG and Gait3D datasets, we set $N_{epoch} = 32$, and the total training epoch is 100 ($\sim$18h).
The AttID encoder $\mathbf{E}$ and video discriminator $\mathbf{D}_{vid}$ are optimized using Adam~\cite{kingma2014adam}, with learning rates of 1e-4 and 2e-4, respectively.

\begin{table}[t]
\centering
\caption{The tuning details of different experiments. LR is the initial learning rate. If LR=1e-4, the initial learning rate of the first two convolution blocks, the last two ones, and the head layers are respectively set to LR, LR×10, and LR×100. $Exp_1$ and $Exp_2$ repressively denote the evaluate of different datasets and gait models.}
\label{tab:finetuning_setting}
\resizebox{\linewidth}{!}{%
\centering
\begin{tabular}{@{}c|c|c|cccc@{}}
\toprule
Experiments                                           & Datasets              & Gait models               & Milestones / Steps    & LR     & \multicolumn{1}{l}{$\xi$} & \multicolumn{1}{l}{$\beta$} \\ \midrule
\multirow{3}{*}{$Exp_1$}    & CCPG                  & \multirow{3}{*}{GaitBase} & (10K, 15K) / 20K      & 1e-4 & 4                         & 1                           \\
                                                     & OU-MVLP               &                           & (10K, 15K) / 20K      & 1e-4 & 4                         & 1                           \\
                                                     & Gait3D                &                           & (10K, 15K) / 20K      & 1e-4 & 4                         & 0.8                         \\ \midrule
\multirow{4}{*}{$Exp_2$} & \multirow{4}{*}{CCPG} & GaitSet                   & (10K, 20K, 30K) / 40K & 1e-1    & 4                         & 1                           \\
                                                     &                       & GaitPart                  & (10K, 15K) / 20K      & 1e-4 & 4                         & 1                           \\
                                                     &                       & GaitBase                  & (10K, 15K) / 20K      & 1e-4 & 4                         & 1                           \\
                                                     &                       & DeepGaitv2                & (20K, 40K, 50K) / 60K      & 1e-1 & 8                         & 2                           \\ \bottomrule
\end{tabular}%
}
\end{table}

\subsection{The Evaluation on Attribute Editing}
\label{sec:user_study}

In this section, we evaluate identity consistency and attribute correctness when editing different attributes by cosine similarity and a user study, respectively.

\textbf{Evaluation of identity consistency.}
The experimental evaluation of identity consistency conducted on the CCPG~\cite{li2023depth} and OU-MVLP dataset~\cite{takemura2018multi} and utilizing the GaitBase model~\cite{fan2023opengait} reveals varied ID-Similarity outcomes when editing different attributes, as shown in Table~\ref{tab:identity_consistency}.
It is worth mentioning that each attribute was edited exactly once with the edited strength set to 1, \textit{i.e.} $\xi = \beta =1$.
The observed identity consistency across various attributes mostly exceeds the median ID-Similarity benchmarks of $0.9296$ ($(0.8806 + 0.9785) / 2$) for the CCPG dataset and $0.7245$ ($(0.9336 + 0.5153) / 2$) for the OU-MVLP dataset, respectively. 
This indicates that editing these attributes by GaitEditor is effective in preserving identity consistency.
The attribute 'Viewpoints' exhibits an unexpectedly high identity consistency score, achieving 0.9380 in the CCPG dataset and 0.7813 in the OU-MVLP dataset.
This is particularly noteworthy considering it involves significant modifications to the body's appearance, suggesting that our proposed GaitEditor can successfully handle the challenge of viewpoint translation.
For the CCPG dataset, the attributes of 'Pants' and 'Hoodie' show marginally lower identity consistency scores. This can be attributed to the significant morphological changes these attributes induce in the leg and neck areas, respectively. 
Such editing likely perturbs the distinctive gait patterns captured by the GaitBase model, which is notably sensitive to editing in these specific regions when trained on the CCPG dataset.



\begin{table}[t]
\centering
\caption{The recognition accuracy of different datasets for GaitBase as the baseline. The rank-1 accuracy percentages presented for each condition demonstrate the performance of the gait models in handling these specific variations.}
\label{tab:gait_recognition_different_datasets}
\resizebox{\linewidth}{!}{%
\centering
\begin{tabular}{@{}c|c|cc@{}}
\toprule
\multirow{2}{*}{Datasets} & \multirow{2}{*}{GaitEditor} & \multicolumn{2}{c}{Rank-1 accuracy (\%)}                                                                     \\ \cmidrule(l){3-4} 
                          &                             & \multicolumn{1}{c|}{Different conditions}                                & \multicolumn{1}{l}{Mean} \\ \midrule
OU-MVLP                   & \ding{56}                   & \multicolumn{1}{c|}{NM: 89.85}                                           & 89.85                    \\
                          & \ding{52}                   & \multicolumn{1}{c|}{\textbf{NM: 90.27}}                                  & \textbf{90.27}           \\ \midrule
\multirow{2}{*}{CCPG}     & \ding{56}                   & \multicolumn{1}{c|}{CL: 71.56, UP: 75.00, DN: 76.85, BG: 78.52}          & 75.48                    \\
                          & \ding{52}                   & \multicolumn{1}{c|}{\textbf{CL: 73.21, UP: 76.98, DN: 77.81, BG: 80.32}} & \textbf{77.08}           \\ \midrule
\multirow{2}{*}{Gait3D}   & \ding{56}                   & \multicolumn{1}{c|}{64.60}                                               & 64.60                    \\
                          & \ding{52}                   & \multicolumn{1}{c|}{\textbf{65.40}}                                      & \textbf{65.40}           \\ \bottomrule
\end{tabular}
}
\end{table}

\begin{table}[t]
\centering
\caption{The recognition accuracy of different gait models on the CCPG dataset. The rank-1 accuracy percentages presented for each condition demonstrate the performance of the gait models in handling these specific variations.}
\label{tab:gait_recognition_different_models}
\resizebox{1\linewidth}{!}{%
\begin{minipage}{1\linewidth}
\centering

\begin{tabular}{c|c|cccc|c} 
\toprule
\multirow{2}{*}{Gait models} & \multirow{2}{*}{GaitEditor} & \multicolumn{5}{c}{Rank-1 accuracy (\%)}                                                                                \\ 
\cmidrule{3-7}
                             &                             & CL                      & UP                      & DN                      & BG                      & Mean            \\ 
\hline
\multirow{2}{*}{GaitSet}     & \ding{56}  & 60.18                   & 65.22                   & 65.11                   & 68.53                   & 64.76           \\
                             & \ding{52}  & \textbf{61.30}          & \textbf{\textbf{66.09}} & \textbf{\textbf{66.21}} & \textbf{\textbf{69.45}} & \textbf{65.76}  \\ 
\cmidrule{1-7}
\multirow{2}{*}{GaitPart}    & \ding{56}  & 64.25                   & 67.76                   & 68.58                   & 71.68                   & 68.07           \\
                             & \ding{52}  & \textbf{\textbf{65.28}} & \textbf{\textbf{70.15}} & \textbf{\textbf{71.36}} & \textbf{\textbf{73.98}} & \textbf{70.19}  \\ 
\cmidrule{1-7}
\multirow{2}{*}{GaitBase}    & \ding{56}  & 71.56                   & 75.00                   & 76.85                   & 78.52                   & 75.48           \\
                             & \ding{52}  & \textbf{\textbf{73.21}} & \textbf{\textbf{76.98}} & \textbf{\textbf{77.81}} & \textbf{\textbf{80.32}} & \textbf{77.08}  \\ 
\cmidrule{1-7}
\multirow{2}{*}{DeepGaitv2}  & \ding{56}  & 78.37                   & 84.67                   & 80.95                   & 89.45                   & 83.36           \\
                             & \ding{52}  & \textbf{\textbf{79.76}} & \textbf{\textbf{85.70}} & \textbf{\textbf{82.30}} & \textbf{\textbf{90.83}} & \textbf{84.65}  \\
\bottomrule
\end{tabular}

\end{minipage}
}
\end{table}


\textbf{Evaluation of attribute correctness.}
To evaluate the attribute correctness of the gait sequences after editing, we conduct a user study due to the unavailable of fine-grained attribute classifiers specific to gait silhouettes.
This study incorporated 101 participants, stratified into two groups: 52 individuals with expertise in gait research and 49 laypersons.
Each participant was presented with ten randomly generated questions, with the format of the inquiries between single-choice and true/false.
The questions were distributed in a weighted manner, with 70\% being single-choice to evaluate the identification of edited attributes and the remaining 30\% formatted as true/false to validate the correctness of a specific attribute edit.
In the single-choice questions, participants were prompted: ``Identify the altered attribute in the video bordered in purple, relative to the reference video bordered in blue." 
The true/false questions were asked: ``Does the video bordered in purple represent an edited `X' attribute, as compared to the reference video bordered in blue?", where `X' means a specific gait attribute.
Here, the sequence bordered in purple represented the edited sequence, while the blue border denoted the original, unedited gait sequence.

The study results, as presented in Table \ref{tab:user_study}, offer a comprehensive view of the correctness associated with various attributes as perceived by subjects of different levels of expertise.
The participants with gait experience exhibited an overall accuracy of 81.46\%, and the non-expert group demonstrated a slightly lower overall accuracy of 71.23\%.
The difference in accuracy is particularly pronounced when it comes to recognizing gender-like (`Masculine' and `Feminine') and age-like (`Younger' and `Older') attributes.
For `Masculine', `Feminine', `Younger' and `Older' subjects without gait experience exhibited decreases -29.42\%, -14.09\%, -11.02\%, and -20.73\% compared to those with gait experience, respectively.
These results reveal that the subtlety of changes in specific regions for gait attributes relies on specialized knowledge for accurate identification.

\begin{table}[t]
\centering
\caption{The anonymization evaluation on different datasets for GaitBase as the baseline. }
\label{tab:gait_ano_different_datasets}
\resizebox{\linewidth}{!}{%
\centering
\begin{tabular}{@{}c|c|c|c@{}}
\toprule
Datasets                 & \begin{tabular}[c]{@{}c@{}}Anonymization \\ methods\end{tabular} & Rank-1 (\%)                                         & FID ($\downarrow$) \\ \midrule
\multirow{3}{*}{OU-MVLP} & Raw                                                              & NM: 89.85                                           & -               \\
                         & Blur                                                             & NM: 11.79                                           & 7.92               \\
                         & \textbf{GaitEditor}                                              & \textbf{NM: 0.61}                                   & \textbf{0.60}      \\ \midrule
\multirow{3}{*}{CCPG}    & Raw                                                              & CL: 71.56, UP: 75.00, DN: 76.85, BG: 78.52          & -               \\
                         & Blur                                                             & CL: 31.84, UP: 27.43, DN: 30.51, BG: 31.65          & 8.36               \\
                         & \textbf{GaitEditor}                                              & \textbf{CL: 13.79, UP: 22.09, DN: 23.93, BG: 12.76} & \textbf{0.56}      \\ \midrule
\multirow{3}{*}{Gait3D}  & Raw                                                              & 64.60                                               & -               \\
                         & Blur                                                             & 20.71                                               & 10.46              \\
                         & \textbf{GaitEditor}                                              & \textbf{8.60}                                       & \textbf{1.05}      \\ \bottomrule
\end{tabular}%
}
\end{table}

\subsection{The Evaluation on Gait Recognition}
The evaluation consists of two aspects: using GaitEditor to enhance different gait models on various gait datasets.

\textbf{Evaluation on different gait datasets.} For comparative analysis, we utilize three distinct gait datasets: the cross-clothing dataset CCPG, the cross-view dataset OU-MVLP, and the outdoor dataset Gait3D, following their officially provided dataset division protocols. 
As baseline models, we employ GaitBase models, officially pre-trained on each of these datasets (CCPG, OU-MVLP, and Gait3D). 
To enhance the performance of the GaitBase model on different datasets, we fine-tune GaitBase using edited input sequences generated by GaitEditor with a probability of 0.2, and the probability of editing viewpoint and non-viewpoint are 0.1 and 0.9, respectively.
The specifics of this fine-tuning process are detailed in Table~\ref{tab:finetuning_setting}.
It is worth mentioning that during the tuning phase of GaitBase when incorporating GaitEditor, its standard data augmentation strategy is involved, adhering to the official training protocols.

Performance improvements achieved through the application of GaitEditor are present in Table~\ref{tab:gait_recognition_different_datasets}. 
Specifically, there are increments of +0.42\%, +1.60\%, and +0.80\% in average rank-1 accuracy over the GaitBase baselines on the OU-MVLP, CCPG, and Gait3D datasets, respectively. 
These enhancements suggest that GaitEditor contributes to more effective learning of gait representations under all cross-clothing, cross-viewpoint, and outdoor conditions.

\textbf{Evaluation on different gait models.} To verify that GaitEditor can enhance various gait methods, we employ four popular gait recognition models: GaitSet, GaitPart, GaitBase, and DeepGaitv2, which follow their official settings and have pre-trained on the CCPG dataset.
Then, we leverage GaitEditor and plug it into different models as an online data augmentation module.
Next, we fine-tune these models using edited input sequences generated by GaitEditor with a probability of 0.2, and the probability of editing viewpoint and non-viewpoint are 0.1 and 0.9, respectively. 
The detailed tuning settings are presented in Table~\ref{tab:finetuning_setting}.

The performance of different gait models on the CCPG dataset using GaitEditor is shown in Table~\ref{tab:gait_recognition_different_datasets}.
The pioneer of the deep-based gait model, GaitSet, when enhanced with GaitEditor, shows an improvement in all conditions, with rank-1 accuracy increasing +1.12\% for CL, +0.87\% for UP, +1.10\% for DN, and +0.92\% for BG, resulting in an average increase +1.00\%. 
Similarly, GaitPart, a method for considering temporal dependencies, also benefits from GaitEditor, with noticeable improvements, particularly in the UP and BG conditions.
DeepGaitv2, a more deep model containing 3D convolution layers, also displays enhanced accuracy with GaitEditor.
These results collectively indicate that the application of GaitEditor for different gait models can effectively enhance the learning of gait representation.

\begin{table}[t]
\centering
\caption{The anonymization evaluation on different gait models on the CCPG dataset. The term 'GaitBase*' denotes the GaitBase model that has been tuned using GaitEditor, following the identical settings outlined in Table~\ref{tab:finetuning_setting}.}
\label{tab:gait_ano_different_models}
\resizebox{\linewidth}{!}{%
\begin{minipage}{\linewidth}
\centering

\begin{tabular}{c|c|cccc|c} 
\toprule
\multirow{2}{*}{Models} & \multirow{2}{*}{GaitEditor} & \multicolumn{5}{c}{Anonymized Rank-1 accuracy ($\downarrow$)}  \\ 
\cmidrule{3-7}
                             &                             & CL    & UP    & DN    & BG    & Mean      \\ 
\cmidrule{1-7}
\multirow{2}{*}{GaitSet}     & \ding{56}  & 60.18 & 65.22 & 65.11 & 68.53 & 64.76     \\
                             & \ding{52}  & \textbf{14.53} & \textbf{19.66} & \textbf{21.31} & \textbf{11.10} & \textbf{16.65}     \\ 
\cmidrule{1-7}
\multirow{2}{*}{GaitPart}    & \ding{56}  & 64.25 & 67.76 & 68.58 & 71.68 & 68.07     \\
                             & \ding{52}  & \textbf{16.96} & \textbf{21.84} & \textbf{23.23} & \textbf{13.20} & \textbf{18.81}     \\ 
\cmidrule{1-7}
\multirow{2}{*}{GaitBase}    & \ding{56}  & 71.56 & 75.00 & 76.85 & 78.52 & 75.48     \\
                             & \ding{52}  & \textbf{13.79} & \textbf{22.09} & \textbf{23.93} & \textbf{12.76} & \textbf{18.14}     \\ 
\cmidrule{1-7}
\multirow{2}{*}{DeepGaitv2}  & \ding{56}  & 71.22 & 76.17 & 75.66 & 80.09 & 75.79     \\
                             & \ding{52}  & \textbf{13.17} & \textbf{23.79} & \textbf{25.48} & \textbf{14.35} & \textbf{14.20}     \\ 
\cmidrule{1-7}
\multirow{2}{*}{GaitBase*}   & \ding{56}  & 73.21 & 76.98 & 77.81 & 80.32 & 77.08     \\
                             & \ding{52}  & \textbf{16.09} & \textbf{26.23} & \textbf{27.62} & \textbf{13.10} & \textbf{20.76}     \\
\bottomrule
\end{tabular}

\end{minipage}
}
\end{table}


\begin{figure*}[t]
    \centering
    \includegraphics[width=\linewidth]{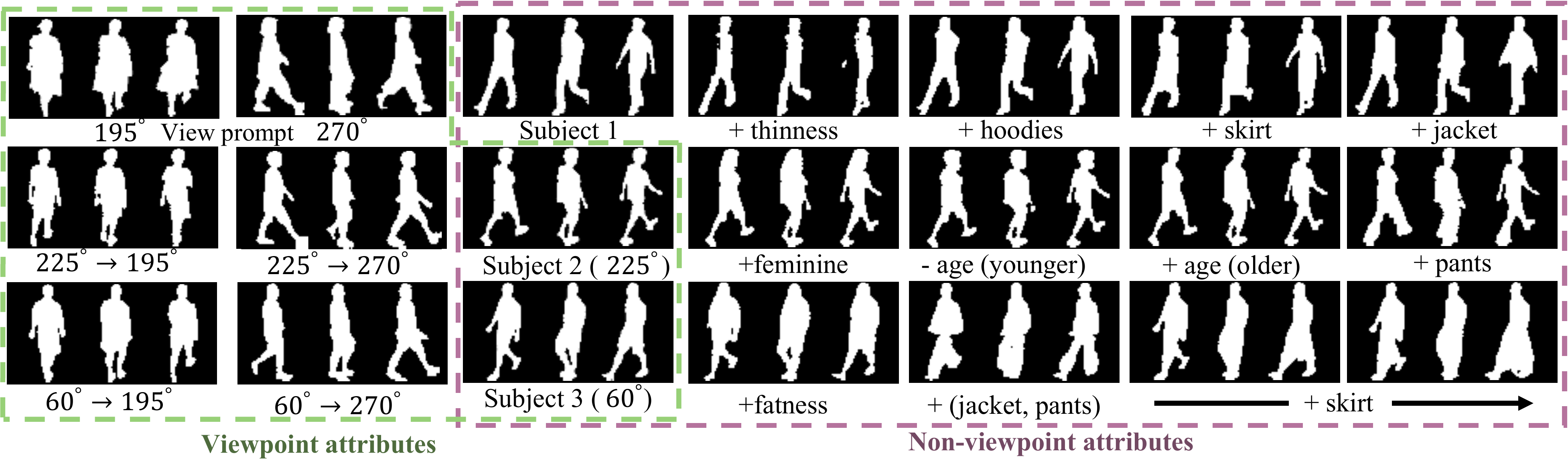}
    \caption{Visualization of gait attribute editing with Non-viewpoint (purple box) and viewpoint (green box) attributes. The $+/-$ means adding/eliminating the attribute, and $a^\circ \rightarrow b^\circ$ means the viewpoint change from $a^\circ$ to $b^\circ$.} 
    \label{fig:fig7_vis_sil}
\end{figure*}

\begin{figure*}[t]
    \centering
    \includegraphics[width=\linewidth]{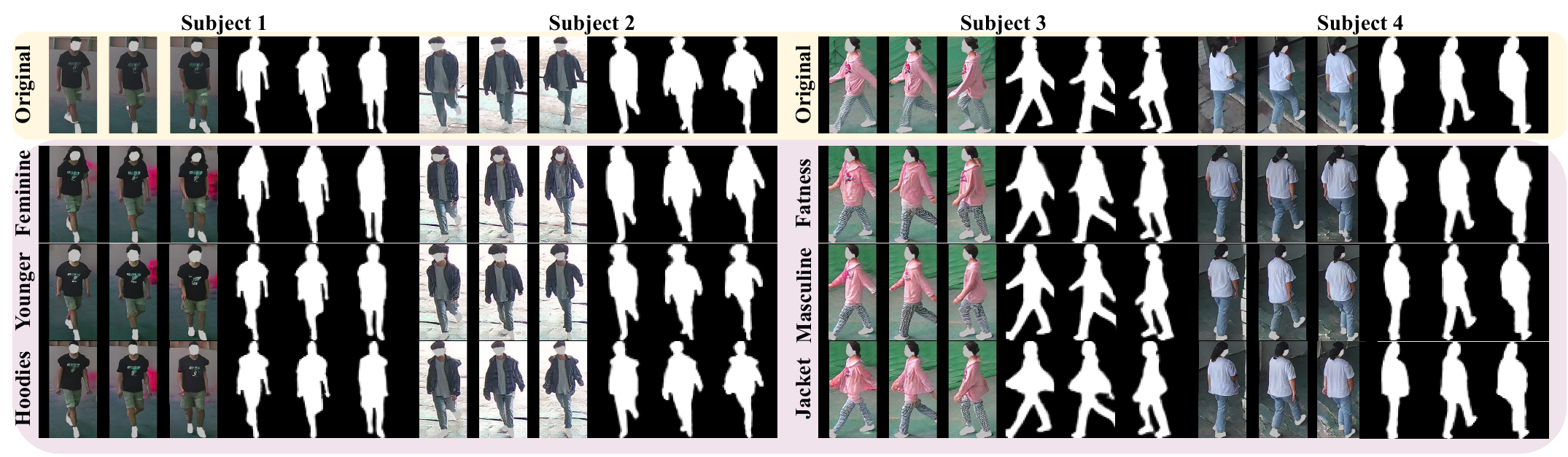}
    \caption{The RGB results of visualization for non-viewpoint attributes. The first row is the original sequences (yellow), and the last three rows are the edited results (purple).} 
    \label{fig:fig8_vis_rgb}
\end{figure*}

\begin{table*}[t]
\centering
\caption{The ablation studies for different modules or loss functions in GaitEditor. The efficacy of these components is quantified using a suite of GAN metrics, which measure the inversion quality, silhouette quality, and identity similarity. The `$\uparrow$' indicates that higher values correspond to improved performance, whereas the `$\downarrow$' means that lower values correspond to improved performance.}
\label{tab:ab_study}
\resizebox{0.7\linewidth}{!}{%
\centering
\begin{tabular}{@{}ccccccc@{}}
\toprule
Settings                         & PSNR ($\uparrow$) & SSIM ($\uparrow$) & FID ($\downarrow$) & LPIPS ($\downarrow$) & MSE ($\downarrow$) & ID-Simility ($\uparrow$) \\ \midrule
\textit{w/o} $\mathcal{L}_{adv}$ & 20.904            & 0.865             & 0.082              & 0.046                & 0.045              & 0.966                    \\
\textit{w/o} $\mathcal{L}_{id}$  & 20.776            & 0.874             & 0.099              & 0.047                & 0.046              & 0.956                    \\
\textit{w/o} $\mathbf{E}_{id}$  & 20.514            & \textbf{0.901}             & \textbf{0.059}              & 0.045                & 0.039              & 0.951                    \\
\textit{w/o} $\mathbf{E}_{mix}$  & 19.884            & 0.848             & 0.154              & 0.054                & 0.060              & 0.969                    \\
Our full model                   & \textbf{21.523}   & 0.902    & 0.069     & \textbf{0.034}       & \textbf{0.030}     & \textbf{0.978}           \\ \bottomrule
\end{tabular}%
}
\end{table*}

\subsection{The Evaluation on Gait Anonymization}
To evaluate the effectiveness of gait anonymization, we use GaitEditor to edit the gait sequence from the entire test probe set and then compute the rank-1 accuracy.
The FID score is computed to evaluate the image quality after editing.
For the evaluation on different datasets, the rank-1 accuracy is evaluated using the GaitBase model, which is pre-trained on the respective training sets of three datasets (CCPG, OU-MVLP and Gait3D).
For the evaluation on different gait models, we employ four popular gait models (GaitSet, GaitPart, GaitBase, and DeepGaitv2), which are pre-trained on the CCPG dataset.

Table~\ref{tab:gait_ano_different_datasets} show the results of the evaluation on different datasets.
Blur involves applying a Gaussian kernel to blur the identity within gait sequences, ensuring anonymity.
Notably, for the cross-view dataset OU-MVLP, the application of GaitEditor plummeted the rank-1 accuracy from 89.848\% to a mere 0.609\%, showcasing a powerful performance in anonymization compared with using blur.
The FID score for silhouette quality also suggests a favourable outcome, with a low value of 0.6038 indicating high-quality image generation post-editing.
Similarly, for the cross-clothing dataset, CCPG, a notable decrease in rank-1 accuracy across various conditions (CL, UP, DN, BG) when GaitEditor is applied.
In the case of the outdoor dataset, Gait3D, the rank-1 accuracy shows a drastic reduction from 64.60\% without GaitEditor to 8.60\% with its application.
In addition, GaitEditor demonstrates a significant superiority over the application of blurring techniques, in terms of both the efficacy of identity anonymization and the quality of images post-anonymization.

Table~\ref{tab:gait_ano_different_models} shows the evaluation results of different gait models tested on the CCPG dataset, with the integration of GaitEditor.
The average rank-1 accuracy of GaitSet, GaitPart, GaitBase, and DeepGaitv2 models shows a substantial decline of -48.11\%, -49.26\%, -57.34\%, and -58.88\%, respectively.
These results demonstrate GaitEditor's efficacy in improving the anonymity of gait data across different model architectures, thus addressing privacy concerns in gait.
Furthermore, the experiment includes an evaluation on the tuned GaitBase model, referred to as GaitBase*.
Despite its optimization through GaitEditor, the diversity of editing in the anonymization process ensures the effectiveness of de-identification, thus significantly facilitating the safeguarding of individual privacy.

\subsection{Visualization}
Fig.~\ref{fig:fig7_vis_sil} shows the faithful gait silhouettes edited by our GaitEditor with various attributes. 
The example gait sequences are from the OU-MVLP dataset.
GaitEditor enables continuous and stackable modification of gait attributes, as shown in the last row of the purple box, via adjusting the strength $\beta$ and editing frequency $\xi$, respectively. 
For viewpoint attributes, GaitEditor is also able to handle the viewpoint translation via two given sequences, and the results are shown in the green box.
Fig.~\ref{fig:fig8_vis_rgb} show the gait images edited by our GaitEditor with non-viewpoint attributes and rendered to RGB format by VQGAN.
The example gait sequences are from the test set of the CCPG dataset.
We can see that GaitEditor can manipulate specific features while maintaining the naturalness of gait.
Compared to previous method~\cite{fu2022stylegan} that relied on training with RGB images, which often focused on editing texture features such as the length and style of clothing, GaitEditor is able to accurately capture and edit attributes related to a pedestrian's body shape, such as the femininity, and body size. 
This distinction highlights the unique ability of GaitEditor to maintain pedestrian privacy. 
It effectively anonymizes identifiable characteristics of individuals without compromising the natural appearance information.

\subsection{Ablation Study}
In this section, we perform an ablation study to evaluate our GaitEditor on the CCPG dataset.
We develop three variants with the modification of the modules and the loss functions:
1) \textit{w/o} $\mathcal{L}_{adv}$, by removing video adversarial loss during training phase.
2) \textit{w/o} $\mathcal{L}_{id}$, by removing identity loss function during training phase.
3) \textit{w/o} $\mathbf{E}_{id}$, by removing the pre-trained gait model and then starting its training from scratch.
4) \textit{w/o} $\mathbf{E}_{mix}$, by removing the feature mix block $\mathbf{E}_{mix}$, then leveraging add operation to fuse different features in latent space.
The quantitative comparisons with various variants can be seen in Table~\ref{tab:ab_study}.
The findings, as shown in Table~\ref{tab:ab_study}, demonstrate that the elimination of either the identity loss term ($\mathcal{L}_{id}$) or the pre-trained gait model ($\mathbf{E}_{id}$) used as prior knowledge results in a reduced ability to preserve identity, as indicated by the ID-Similarity metric.
It's worth noting that variant $\mathbf{E}_{id}$ has the best performance on metrics SSIM and FID but has the worst performance on metric ID-Simility.
This is because the pre-trained gait model ensures the consistency of identity in the inverted frames. However, this also slightly compromises the quality of the intra-frame imaging.
Omitting the adversarial loss results in a lower inversion quality, which is evident from the decreased PSNR and increased FID scores.
Notably, the absence of the feature mix block $\mathbf{E}_{mix}$ has a pronounced impact, leading to a substantial drop in image quality.
Our full model, which integrates all components, achieves the highest scores across all metrics. 





\section{Limitations}
We acknowledge that our method has some limitations, such as 1) difficulties in accurately translating viewpoints for substantial changes, for instance, achieving a $180^\circ$ rotation; 2) obstacles in accurately editing attributes of age-like and gender-like, especially when the edits present contradictions to conventional perceptions, like a man having long hair; 3) challenges associated with editing attributes in the temporal dimension, such as the walking phase.

\section{Conclusion and future work}\label{sec6}

We presented GaitEditor, a pioneering gait attribute editing methodology for editing various gait attributes with applications in gait recognition and anonymization. 
The motivation for GaitEditor stemmed from the challenge of creating a unified framework that addresses the dual objectives of enabling improved individual identification while mitigating privacy concerns. 
GaitEditor's efficacy was quantitatively evaluated on three widely used gait benchmarks and against four renowned gait recognition algorithms, demonstrating its versatility and effectiveness in both recognition and anonymization tasks.
Additionally, its capability to generate vivid silhouettes and RGB sequences further underscores its application in gait recognition and anonymization.

As we look ahead, learning the general gait representation from unlabelled walking videos is a challenging yet highly appealing task, as collecting a large amount of annotated gait data is costly and insatiable. 
GaitEditor provides a potential solution by serving as a data augment module, thereby benefiting \textbf{unsupervised gait recognition}.
Meanwhile, GaitEditor also hopefully benefits \textbf{Cross-X gait recognition}. 
Cross-view and cross-clothing are two of the most prominent testing setups for current gait datasets, and as a result, most existing methods are largely oriented toward these issues. However, in real applications, the recognition model often encounters many other complex situations, such as the cross-hat, cross-hair, cross-size, cross-age, and even more specialized, cross-X cases. Due to the lack of fine-grained labelled data pairs, it is difficult to accurately evaluate cross-X-oriented gait recognition. Fortunately, GaitEditor provides a potential solution by enabling the simulation of virtual datasets with a wide variety of gait attribute changes.
Last but not least, although GaitEditor can anonymize the identity within gait sequences by perturbing body appearance, the body appearance also has unnatural results for display. 
For instance, family members are more willing to see the real face and body size rather than the anonymized one, and people may want to share data that only certain authorized parties can interpret.
Therefore, the \textbf{reversibility of the gait anonymization} is also needed.

In summary, GaitEditor represents a pivotal advancement in our ability to edit multiple gait attributes simultaneously through a unified framework. 
It showcases potential capabilities for boosting the scale and diversity of real datasets, thereby expected to inspire grand ideas that free the gait applications from the restrictions imposed by real datasets in terms of scale and diversity.

\bibliographystyle{IEEEtran}
\bibliography{tpami_main}

\vfill\eject

\end{document}